%% file: neurips_2021.tex
\documentclass{article}

\PassOptionsToPackage{numbers, compress}{natbib}

    \usepackage[final]{neurips_2021}

\usepackage[utf8]{inputenc} %
\usepackage[T1]{fontenc}    %
\usepackage[colorlinks=true]{hyperref}       %
\usepackage{url}            %
\usepackage{booktabs}       %
\usepackage{amsfonts}       %
\usepackage{nicefrac}       %
\usepackage{microtype}      %
\usepackage{xcolor}         %
\usepackage{graphicx}
\usepackage{subfigure}
\usepackage{amsthm}
\usepackage{mathtools}
\usepackage{multirow}
\usepackage{float}
\usepackage{array}
\usepackage{pifont}
\usepackage{wrapfig}
\usepackage{algorithm}
\usepackage{algorithmic}
\usepackage{etoc}

\input{math_commands}

\usepackage{etoc}
\usepackage{lipsum}
\usepackage[textsize=scriptsize]{todonotes}
\newtheorem{myprop}{\bf{Proposition}}

\newtheorem{mythr}{\bf{Theorem}}
\newtheorem{mylemma}{\bf{Lemma}}

\DeclarePairedDelimiterX{\inp}[2]{\langle}{\rangle}{#1, #2}

\DeclareMathOperator*{\minimize}{\text{minimize}}
\DeclareMathOperator*{\maximize}{\text{maximize}}

\DeclareMathOperator*{\st}{\text{subject to}}
\DeclareMathAlphabet\mathbfcal{OMS}{cmsy}{b}{n}
\newcommand{\Def}[0]{\mathrel{\mathop:}=}

\usepackage{adjustbox}
\usepackage{comment}
\usepackage{hyperref}

\title{Adversarial Attack Generation Empowered by Min-Max Optimization}

\author{
Jingkang Wang$^{1,2}$\thanks{Equal contributions.} \quad Tianyun Zhang$^{3*}$\quad Sijia Liu$^{4,5}$ \vspace{0.6mm} \\ \vspace{0.6mm} \textbf{Pin-Yu Chen}$^{5}$ \quad \textbf{Jiacen Xu}$^{6}$ \quad \textbf{Makan Fardad}$^{7}$ \quad \textbf{Bo Li}$^{8}$ \\
University of Toronto$^{1}$, Vector Institute$^{2}$, Cleveland State University$^{3}$ \\
Michigan State University$^{4}$, MIT-IBM Watson AI Lab, IBM Research$^{5}$ \\ University of California, Irvine$^{6}$, Syracuse University$^{7}$ \\ University of Illinois at Urbana-Champaign$^{8}$ 
}

\begin{document}

\maketitle

\etocdepthtag.toc{mtchapter}
\etocsettagdepth{mtchapter}{none}
\etocsettagdepth{mtappendix}{none}

\begin{abstract}
The worst-case training principle that minimizes the maximal adversarial loss, also known as adversarial training (AT), has shown to be a state-of-the-art approach for enhancing adversarial robustness.
Nevertheless, min-max optimization beyond the purpose of AT has not been rigorously explored in the adversarial context. In this paper, we show how a general framework of min-max optimization over multiple domains can be leveraged to advance the design of different types of adversarial attacks. In particular, given a set of risk sources, minimizing the worst-case attack loss can be reformulated as a min-max problem by introducing  domain weights that are maximized over the probability simplex of the domain set. We showcase this unified framework in three attack generation problems -- attacking model ensembles, devising universal perturbation under multiple inputs, and crafting attacks resilient to data transformations. 
 Extensive experiments demonstrate that our approach leads to substantial attack improvement over the existing heuristic strategies as well as robustness improvement over state-of-the-art defense methods trained to be robust against multiple perturbation types. 
Furthermore, we find that the self-adjusted domain weights learned from our min-max framework can provide a holistic tool to explain the difficulty level of attack across domains. Code is available at \url{https://github.com/wangjksjtu/minmax-adv}.
\end{abstract}

\section{Introduction}
Training a machine learning model that is capable of assuring its worst-case performance against possible adversaries given a specified threat model is a fundamental and challenging problem, especially for deep neural networks (DNNs) \cite{szegedy2013intriguing,Goodfellow2015explaining,carlini2017towards,xiao2018generating,xiao2018spatially}. A common practice to train an adversarially robust model is based on a specific form of min-max training, known as \textit{adversarial training} (AT)~\cite{Goodfellow2015explaining,madry2017towards}, where the minimization step learns model weights under the adversarial loss constructed at the maximization step in an alternative training fashion.  
In practice, AT has achieved the state-of-the-art defense performance against $\ell_p$-norm-ball input perturbations \cite{athalye2018obfuscated}.

Although the  min-max principle is widely used in AT and its variants \cite{madry2017towards,sinha2018certifying,zhang2019theoretically,tramer2019adversarial}, few work has studied its power in attack generation.
Thus, we ask:
\textit{Beyond AT, can other types of min-max formulation and optimization techniques advance the research in adversarial attack generation?} 
In this paper, we give an affirmative answer corroborated by the substantial performance gain and the ability of self-learned risk interpretation using our proposed min-max framework on several tasks for adversarial attack. 

We demonstrate the utility of a general formulation for minimizing the maximal loss induced from a set of risk sources (domains). Our considered min-max formulation is fundamentally different from AT, as our maximization step is taken over the probability simplex of the set of domains. Moreover, we show that many problem setups in adversarial attacks can in fact be reformulated under this general min-max framework, including attacking model ensembles \cite{tram2018ensemble,caad_ensemble}, devising universal perturbation to input samples \cite{moosavi2016universal} and data transformations \cite{athalye2017synthesizing,brown2017}. %
However, current methods for solving these tasks often rely on simple heuristics (e.g., uniform averaging), resulting in significant performance drops when comparing to our proposed min-max optimization framework.

\paragraph{Contributions}
\ding{172} With the aid of min-max optimization, 
we propose a unified
\underline{{a}}lternating  one-step \underline{{p}}rojected \underline{{g}}radient \underline{{d}}escent-\underline{{a}}scent  (APGDA) attack method, which can readily be specified to generate model ensemble attack, universal attack over multiple images, and robust attack over data transformations. \ding{173} In theory, we show that APGDA has an $O(1/T)$ convergence rate, where $T$ is the number of iterations. In practice, we show that APGDA obtains 17.48\%, 35.21\% and 9.39\% improvement on average compared with conventional min-only PGD attack methods on CIFAR-10.
\ding{174} More importantly, we demonstrate that by tracking the learnable weighting factors associated with multiple domains, our method can provide tools for self-adjusted importance assessment on the mixed learning tasks.
\ding{175} Finally, we adapt the idea of the domain weights into a defense setting~\cite{tramer2019adversarial}, where multiple $\ell_p$-norm perturbations are generated, and achieve superior performance as well as intepretability.

\subsection{Related work}
Recent studies have identified that DNNs are highly vulnerable to adversarial manipulations in various applications~\cite{szegedy2013intriguing,carlini2016hidden,Jia2017adv,Lin2017adv,huang2017adv,carlini2018audio,zhao2018adv,eykholt2018robust,chen2018attacking,lei2018discrete}, thus leading to an arms race between adversarial attacks~\cite{carlini2017towards,athalye2018obfuscated,goodfellow2014explaining,papernot2016limitations,moosavi2016deepfool,xu2018structured,square_attack,autoattack} and defenses~\cite{madry2017towards,sinha2018certifying,zhang2019theoretically,tramer2019adversarial,meng2017magnet,xie2017mitigating,xu2018feature,WongRK20,overfit_at,CohenRK19}. One intriguing property of adversarial examples is the transferability across multiple domains~\cite{liu2016delving,tramer2017space,papernot2017practical,su2018robustness}, which indicates a more challenging yet promising research direction -- devising universal adversarial perturbations over model ensembles~\cite{tram2018ensemble,caad_ensemble}, input samples~\cite{moosavi2016universal,metzen2017uni,universal_AT} and data transformations~\cite{athalye2018obfuscated,athalye2017synthesizing,brown2017}. %

Besides, many recent works started to produce physical realizable perturbations that expose real world threats. %
The most popular approach~\cite{AthalyeEIK18,EykholtEF0RXPKS18}, as known as Expectation Over Transformation (EOT), is to train the attack under different data transformation (e.g., different view angles and distances).
However, current approaches suffer from a significant performance loss for resting on the uniform averaging strategy or heuristic weighting schemes~\cite{caad_ensemble,universal_AT}. We will compare these works with our min-max method in Sec.~\ref{sec:experiments}. As a natural extension following min-max attack, we study the generalized AT under multiple perturbations     ~\cite{tramer2019adversarial,araujo2019robust, kang2019testing,croce2019provable}. Finally, our min-max framework is adapted and inspired by previous literature on robust optimization over multiple domains~\cite{qianzhu18,rafique2018non,lutsahong18,lu2019understand}.

To our best knowledge, only few works leverage min-max principle for adversarial attack generation while the idea of producing the worst-case example across multiple domains is quite natural. Specifically, \cite{bose20} considered the non-interactive blackbox adversary setting and proposed a framework that models the crafting of adversarial examples as a min-max game between a generator of attacks and a classifier.
\cite{SheikholeslamiL21} introduced a min-max based adaptive attacker’s objective to craft perturbation so that it simultaneously evades detection and causes misclassification.
Inspired by our work, the min-max formulation has also been extended to zero-order blackbox attacks~\cite{liu19icml} and physically realizable attacks~\cite[Adversarial T-shirt]{XuZ0FSCCWL20}. We hope our unified formulation can stimulate further research on applying min-max principle and interpretable domain weights in more attack generation tasks that involve in evading multiple systems.

\section{Min-Max Across Domains}
Consider $K$ loss functions $\{ F_i(\mathbf v)\}$ (each of which is defined on a learning domain), the problem of robust  learning  over $K$ domains can be formulated as  \cite{qianzhu18,rafique2018non,lutsahong18} %
\begin{align}\label{eq: prob0}
    \begin{array}{ll}
\displaystyle \minimize_{\mathbf v \in \mathcal V} \, \maximize_{\mathbf w \in \mathcal P}          &  \sum_{i=1}^K w_i F_i( \mathbf v ),
    \end{array}
\end{align}
where $ \mathbf v $ and $\mathbf w$ are optimization variables,  
$\mathcal V$ is a   constraint set, and $\mathcal P$ denotes the probability simplex $\mathcal P = \{ \mathbf w \, | \,  \mathbf 1^T \mathbf w = 1, w_i \in [0,1], \forall i \}$.
Since the inner maximization  problem in \eqref{eq: prob0} is a linear function of $\mathbf w$ over the  probabilistic simplex, problem \eqref{eq: prob0} is thus equivalent to 
\begin{align}\label{eq: prob0_1}
        \begin{array}{ll}
\displaystyle \minimize_{\mathbf v \in \mathcal V} \, \maximize_{i \in [K]}          &  F_i (\mathbf {v}),
    \end{array}
\end{align}
where $[K]$ denotes the integer set $\{ 1,2,\ldots, K\}$.
   
\paragraph{Benefit and   Challenge from  \eqref{eq: prob0}.}
Compared to  multi-task learning in a {finite-sum} formulation which minimizes $K$ losses on \textit{average},  problem \eqref{eq: prob0} provides %
consistently robust \textit{worst-case} performance across all domains. This can be explained from the epigraph form of  \eqref{eq: prob0_1},
\begin{align}\label{eq: prob0_1_epi}
    \begin{array}{ll}
\displaystyle \minimize_{\mathbf v \in \mathcal V, t} ~ t,
~~~\st ~  F_i( \mathbf v ) \leq t, i \in [K],
    \end{array}
\end{align}
where $t$ is an epigraph variable \cite{boyd2004convex} that provides the $t$-level robustness at each domain.

In computation,
the inner maximization  problem of \eqref{eq: prob0} always returns the one-hot value of $\mathbf w$, namely,
$\mathbf w = \mathbf e_i$, where $\mathbf e_i$ is the $i$th standard basis vector, and $i = \argmax_i \{ F_i(\mathbf v) \}$.
However, this one-hot coding reduces the generalizability to other domains and
induces instability of the learning procedure in practice. Such an issue is often mitigated by introducing a \textit{strongly concave regularizer} in the inner maximization step to strike a balance between the average and the worst-case performance \cite{lutsahong18,qianzhu18}. 

\paragraph{Regularized Formulation.}
Following \cite{qianzhu18},
we penalize the distance between the \textit{worst-case} loss and the \textit{average} loss over $K$ domains. This yields  
\begin{align}\label{eq: prob0_reg}
  \hspace*{-0.05in}  \begin{array}{ll}
\displaystyle \minimize_{\mathbf v \in \mathcal V} \, \maximize_{\mathbf w \in \mathcal P}          &  \sum_{i=1}^K w_i F_i( \mathbf v ) - \frac{\gamma}{2} \| \mathbf w - \mathbf 1/K \|_2^2,
    \end{array} 
\end{align}
where $\gamma > 0$ is a regularization parameter. %
As $\gamma \to 0$, problem \eqref{eq: prob0_reg} is equivalent to \eqref{eq: prob0}. By contrast,  it becomes the finite-sum problem when $\gamma \to \infty $ since $\mathbf w \to \mathbf 1/ K$. \textit{In this sense, the trainable $\mathbf w$ provides an essential indicator on  the importance level of  each domain.} The larger the weight is, the more important the domain is. We call $\mathbf w$ \textit{domain weights} in this paper.

\section{Min-Max Power in Attack Design}
\label{sec: min_max_atk}
To the best of our knowledge,
few work has  studied the power of min-max in attack generation.
In this section, we demonstrate how the unified min-max framework~\eqref{eq: prob0_reg} fits into various attack settings. With the help of domain weights, our solution yields better empirical performance and explainability. Finally, we present the min-max algorithm with convergence analysis to craft %
robust
perturbations against multiple domains.

\subsection{A Unified Framework for Robust Adversarial Attacks}\label{sec: robust_attack}
The general goal of adversarial attack is to craft an adversarial example $\mathbf x^\prime = \mathbf x_0 + \boldsymbol{\delta} \in  \mathbb R^d$
to mislead the prediction of machine learning (ML) or deep learning (DL) systems, where $\mathbf x_0$ denotes the natural example   with the true label $t_0$, and  $\boldsymbol{\delta}$ is known as \textit{adversarial perturbation}, commonly subject to $\ell_p$-norm ($p \in \{ 0,1,2, \infty  \}$) constraint  
$\mathcal X \Def \{ \boldsymbol{\delta } \, | \, \| \boldsymbol{\delta } \|_p \leq \epsilon, ~  \mathbf x_0 + \boldsymbol{\delta}  \in [0,1]^d \}$ 
for a given small number $\epsilon$. Here   the    $\ell_p$ norm  enforces the similarity between $\mathbf x^\prime$ and $\mathbf x_0$, and the input space of ML/DL systems  is normalized to $[0,1]^d$.

\paragraph{Ensemble Attack over Multiple Models.}
Consider $K$  ML/DL models $\{ \mathcal M_i \}_{i=1}^K$, the goal is to find robust adversarial examples that can  fool all $K$ models \textit{simultaneously}. 
In this case, the notion of `domain' in  \eqref{eq: prob0_reg} is specified  as `model', and
the objective function $F_i$ in \eqref{eq: prob0_reg} signifies  the attack loss $f(\boldsymbol{\delta}; \mathbf x_0, y_0, \mathcal M_i)$ given the natural input $(\mathbf x_0, y_0 )$ and  the model $\mathcal M_i$. Thus, problem \eqref{eq: prob0_reg} becomes
\begin{align}\label{eq: prob0_ensemble}
    \begin{array}{ll}
\displaystyle \minimize_{ \boldsymbol{\delta }\in \mathcal X} \, \maximize_{\mathbf w \in \mathcal P}          &  \sum_{i=1}^K w_i f(\boldsymbol{\delta}; \mathbf x_0, y_0, \mathcal M_i) - \frac{\gamma}{2} \| \mathbf w - \mathbf 1/K \|_2^2,
    \end{array}
\end{align}
where $\mathbf w$ encodes the difficulty level of  attacking each model.

\paragraph{Universal Perturbation over Multiple Examples.}
Consider $K$ natural examples $\{ ( \mathbf x_i, y_i) \}_{i=1}^K$ and a single  model  $\mathcal M$, our goal is to find the universal perturbation $\boldsymbol{\delta}$ so that all the corrupted $K$ examples can fool $\mathcal M$. 
In this case,  the notion of `domain' in  \eqref{eq: prob0_reg} is specified  as `example', and
problem \eqref{eq: prob0_reg} becomes
\begin{align}\label{eq: prob0_uni}
    \begin{array}{ll}
\displaystyle \minimize_{ \boldsymbol{\delta }\in \mathcal X} \, \maximize_{\mathbf w \in \mathcal P}          &  \sum_{i=1}^K w_i f(  \boldsymbol{\delta } ; \mathbf x_i, y_i, \mathcal M) - \frac{\gamma}{2} \| \mathbf w - \mathbf 1/K \|_2^2,
    \end{array}
\end{align}
where different from \eqref{eq: prob0_ensemble},
$\mathbf w$ encodes the difficulty level of attacking each example. 

\paragraph{Adversarial Attack over Data Transformations.}
Consider $K$ categories of data transformation $\{ p_i \}$, e.g.,  rotation, lightening, and translation, our goal is to find the adversarial attack that is robust to data transformations. Such an attack setting is commonly applied to generate physical adversarial examples~\cite{athalye18b,eykholt2018robust}.
Here the notion of `domain' in  \eqref{eq: prob0_reg} is specified  as `data transformer', and
problem \eqref{eq: prob0_reg} becomes
\begin{align}
\hspace{-0.15in}    \begin{array}{ll}
\displaystyle \minimize_{ \boldsymbol{\delta }\in \mathcal X} \, \maximize_{\mathbf w \in \mathcal P}          &  \sum_{i=1}^K w_i \mathbb E_{t \sim p_i }[ f( t( \mathbf x_0 + \boldsymbol{\delta }  ) ;  y_0, \mathcal M ) ] - \frac{\gamma}{2} \| \mathbf w - \mathbf 1/K \|_2^2 ,
    \end{array} \hspace{-0.1in} 
    \label{eq: prob0_phy}
\end{align}
where $\mathbb E_{t \sim p_i }[ f( t( \mathbf x_0 + \boldsymbol{\delta }  ) ;  y_0, \mathcal M ) ]$ denotes the attack loss under the distribution of data transformation $p_i$, and $\mathbf w$  encodes the difficulty level of attacking each type of transformed  example $\mathbf x_0$. We remark that if $\mathbf w = {\mathbf 1}/{K}$, then problem \eqref{eq: prob0_phy} reduces to the existing expectation of transformation (EOT) setup used for physical attack generation \cite{athalye18b}.

\paragraph{{Benefits of Min-Max Attack Generation with Learnable Domain Weights $\mathbf w$:}} 
We can interpret \eqref{eq: prob0_ensemble}-\eqref{eq: prob0_phy} as finding the \textit{robust} adversarial attack against the \textit{worst-case environment}  that an  adversary encounters, e.g., multiple victim models, data samples, and input transformations. The proposed min-max design of adversarial attacks leads to two main benefits.
First, compared to the heuristic weighting strategy (e.g., clipping thresholds on the importance of individual attack losses \cite{universal_AT}), our proposal is free of 
supervised manual adjustment on domain weights. 
Even by carefully tuning the heuristic weighting strategy, we  find that our approach with self-adjusted $\mathbf w$ consistently outperforms the clipping strategy in \cite{universal_AT} (see Table~\ref{tab:ensemble_heuristic}).
Second, the learned domain weights can be used to assess the model robustness when facing different types of adversary. We refer readers to Figure~\ref{fig:mnist_ensemble}c and Figure~\ref{tab:uni_interpret_short} for more details.

\subsection{Min-Max Algorithm for Adversarial Attack Generation}

\begin{wrapfigure}{R}{0.47\textwidth}
\vspace{-7mm}
\begin{small}
\begin{minipage}{0.47\textwidth}
\begin{algorithm}[H]
\caption{APGDA to solve problem \eqref{eq: prob0_reg}}
\begin{algorithmic}[1]
\STATE Input: given $\mathbf w^{(0)}$ and $\boldsymbol{\delta}^{(0)}$.
\FOR{$t =  1,2,\ldots, T$}
\STATE \textit{outer min.}: fixing   $\mathbf w = \mathbf w^{(t-1)}$, call PGD 
\eqref{eq: pgd_out_min} to update  $ \boldsymbol{\delta}^{(t)}$ 
\STATE \textit{inner max.}: fixing $\boldsymbol{\delta} = \boldsymbol{\delta}^{(t)}$,  update $\mathbf w^{(t)}$ with projected gradient ascent \eqref{eq: pgd_in_max}
\ENDFOR  
\end{algorithmic}\label{alg: min_max_general}
\end{algorithm}
\end{minipage}
\end{small}
\vspace{-3mm}
\end{wrapfigure}

We propose the \underline{\textbf{a}}lternating %
\underline{\textbf{p}}rojected \underline{\textbf{g}}radient \underline{\textbf{d}}escent-\underline{\textbf{a}scent}  (APGDA) method (Algorithm\,\ref{alg: min_max_general}) to solve problem \eqref{eq: prob0_reg}.
For ease of presentation, 
we write problems \eqref{eq: prob0_ensemble}, \eqref{eq: prob0_uni}, \eqref{eq: prob0_phy} into the general form
\begin{align} %
{\small
 \hspace*{-0.1in}   \begin{array}{ll}
\displaystyle \minimize_{\boldsymbol{\delta} \in \mathcal X} \, \maximize_{\mathbf w \in \mathcal P}          &  \sum_{i=1}^K w_i F_i( \boldsymbol{\delta} ) - \frac{\gamma}{2} \| \mathbf w - \mathbf 1/K \|_2^2,
    \end{array}\hspace*{-0.05in}
\nonumber
}
\end{align}
\normalsize
where $F_i$ denotes the $i$th individual attack loss.
We show that at each iteration, APGDA takes only one-step PGD for outer minimization and
one-step projected gradient ascent for inner maximization. 

\paragraph{Outer Minimization}
Considering $\mathbf w = \mathbf w^{(t-1)}$  and $ F(\boldsymbol{\delta}) \Def \sum_{i=1}^K w_i^{(t-1)} F_i(\boldsymbol{\delta}) $ in \eqref{eq: prob0_reg}, we perform one-step PGD   to update  $\boldsymbol{\delta}$ at iteration $t$,
\begin{align}\label{eq: pgd_out_min}
\begin{array}{l}
    \boldsymbol{\delta}^{(t)} = \mathrm{proj}_{\mathcal X} \left ( \boldsymbol{\delta}^{(t-1)} - \alpha \nabla_{\boldsymbol{\delta}} F  (\boldsymbol{\delta}^{(t-1)}  ) \right  ),
\end{array}
\end{align}
where $\mathrm{proj}(\cdot)$ denotes the Euclidean projection operator, i.e., $\mathrm{proj}_{\mathcal X} (\mathbf a) = \argmin_{\mathbf x \in \mathcal X} \| \mathbf x - \mathbf a \|_2^2$ at the point $\mathbf a$,  
$\alpha > 0$ is a given learning rate, and $\nabla_{\boldsymbol{\delta}}$ denotes  the first-order gradient w.r.t. $\boldsymbol{\delta}$.
  If $p = \infty$, then the projection function becomes the clip function. 
  In Proposition~\ref{lemma_Euclidean_projection}, we derive the solution of $\mathrm{proj}_{\mathcal X} (\mathbf a) $ under different $\ell_p$ norms for $p \in \{ 0,1,2 \}$.

\begin{myprop}
\label{lemma_Euclidean_projection}
Given a point $\mathbf  a \in \mathbb R^d$ and a constraint set $\mathcal X = \{  \boldsymbol{\delta} | \| \boldsymbol{\delta} \|_p \leq \epsilon, \check {\mathbf c} \leq \boldsymbol{\delta} \leq \hat {\mathbf c}\}$, the Euclidean projection $\boldsymbol{\delta}^* = \mathrm{proj}_{\mathcal X} (\mathbf a) $  has a closed-form solution when $p \in \{ 0,1,2 \}$, where the specific form is given by Appendix\, \ref{app: lemma_Euclidean_projection_ap}.
\end{myprop}

\paragraph{Inner Maximization}
By fixing $\boldsymbol{\delta} = \boldsymbol{\delta}^{(t)}$ and letting 
$\psi(\mathbf w) \Def   \sum_{i=1}^K w_i F_i( \boldsymbol{\delta}^{(t)} ) - \frac{\gamma}{2} \| \mathbf w - \mathbf 1/K \|_2^2$ in problem \eqref{eq: prob0_reg}, we then perform one-step  PGD (w.r.t.  $-\psi$) to update   $\mathbf w$,
\begin{align}\label{eq: pgd_in_max}
    \mathbf w^{(t)}&= \mathrm{proj}_{\mathcal P} \Big( \underbrace{\mathbf w^{(t-1)} +\beta  \nabla_{\mathbf w} \psi(\mathbf w^{(t-1)}) }_\text{ $\mathbf b $ } \Big) = \left ( \mathbf b- \mu \mathbf 1 \right )_+,  
\end{align}%
where $\beta > 0$ is a given learning rate,   $\nabla_{\mathbf w} \psi(\mathbf w) =   \boldsymbol  {\phi}^{(t)} - \gamma (\mathbf w - \mathbf 1/K)$, and $\boldsymbol  {\phi}^{(t)} \Def [F_1(\boldsymbol{\delta}^{(t)}), \ldots,F_K(\boldsymbol{\delta}^{(t)}) ]^T$.
In \eqref{eq: pgd_in_max},  
  the second equality holds due to the closed-form of projection operation onto the 
probabilistic simplex $\mathcal P$ \cite{parikh2014proximal}, where $(x)_+ = \max \{ 0, x\}$, and $\mu $ is the  root of the equation 
$
    \mathbf 1^T (  \mathbf b - \mu \mathbf 1 )_+ = 1.
$
Since $\mathbf 1^T (  \mathbf b  - \min_i \{ b_i  \}  \mathbf 1 + \mathbf 1/K )_+ \geq \mathbf 1^T \mathbf 1/K  = 1$, and
$\mathbf 1^T ( \mathbf b -\max_i \{ b_i \} \mathbf 1 + \mathbf 1/K )_+ \leq \mathbf 1^T \mathbf 1/K = 1 $, the root $\mu$ exists within the interval $[ \min_i \{ b_i  \} - 1/K, \max_i \{ b_i  \} - 1/K ]$ and can be
found via the bisection method \cite{boyd2004convex}.

\paragraph{Convergence Analysis}
We remark that APGDA follows the gradient primal-dual optimization framework \cite{lu2019understand}, and thus enjoys the same optimization guarantees. 

\begin{mythr}%
Suppose that in problem \eqref{eq: prob0_reg}  $F_i( \boldsymbol \delta )$  has $L$-Lipschitz continuous gradients, and $\mathcal X$ is a convex compact set.  %
Given learning rates $\alpha \leq \frac{1}{  L} $ and $\beta < \frac{1}{\gamma}  $, then  the sequence $\{ \boldsymbol{\delta}^{(t)}, \mathbf w^{(t)} \}_{t=1}^T$  generated by Algorithm\,\ref{alg: min_max_general} converges to a first-order  stationary point\footnote{The stationarity is measured by the $\ell_2$ norm of   gradient of the objective in \eqref{eq: prob0_reg} w.r.t. $(\boldsymbol{\delta},\mathbf w)$.} in rate $\mathcal{O}\left(\frac{1}{T}\right)$. 
\label{thr: conv}
\end{mythr}
\textit{Proof}: Note that the objective function of problem \eqref{eq: prob0_reg} is strongly concave w.r.t. $\mathbf w$ with parameter $\gamma$, and has $\gamma$-Lipschitz continuous gradients. Moreover, we have $\| \mathbf w \|_2 \leq 1$ due to $\mathbf w \in \mathcal P$.
Using these facts and Theorem 1 in \cite{lu2019understand} or \cite{lu2019block} completes the proof.
\hfill $\square$

\section{Experiments on Adversarial Exploration}
\label{sec:experiments}
In this section, we first evaluate the  proposed min-max optimization strategy on three attack  %
tasks.
We show that our approach leads to 
substantial improvement 
compared with state-of-the-art attack methods such as average ensemble PGD~\cite{caad_ensemble} and EOT~\cite{athalye2018obfuscated,brown2017,athalye18b}.
We also demonstrate the effectiveness of learnable domain weights in guiding the adversary's exploration over multiple domains.

\subsection{Experimental setup}
We thoroughly evaluate our algorithm on MNIST and CIFAR-10. 
A set of diverse image classifiers (denoted from Model A to Model H) are trained, including multi-layer perceptron (MLP), All-CNNs~\cite{allcnns2015}, LeNet~\cite{Lecun1998gradient}, LeNetV2, VGG16~\cite{vgg16}, ResNet50~\cite{he2016deep}, Wide-ResNet~\cite{madry2017towards,wide_resnet} and GoogLeNet~\cite{googlenet}. 
The details about model architectures and training process are provided in Appendix~\ref{ap:model_arch}. Note that problem formulations \eqref{eq: prob0_ensemble}-\eqref{eq: prob0_phy} are applicable to both \textit{untargeted} and \textit{targeted} attack. Here we focus on the former setting and use C\&W loss function~\cite{carlini2017towards,madry2017towards} with a confidence parameter $\kappa = 50$. The adversarial examples are generated by 20-step PGD/APGDA unless otherwise stated (e.g., 50 steps for ensemble attacks). APGDA algorithm is relatively robust and will not be affected largely by the choices of hyperparameters ($\alpha, \beta, \gamma$). Apart from absolute attack success rate (ASR), we also report the relative improvement or degradationon the worse-case performance in experiments: Lift($\uparrow$).
The details of crafting adversarial examples are available in Appendix~\ref{ap:craft_adv_examples}.

\begin{figure*}[t]
\begin{tabular}{ccc}
    \includegraphics[width=0.31\textwidth]{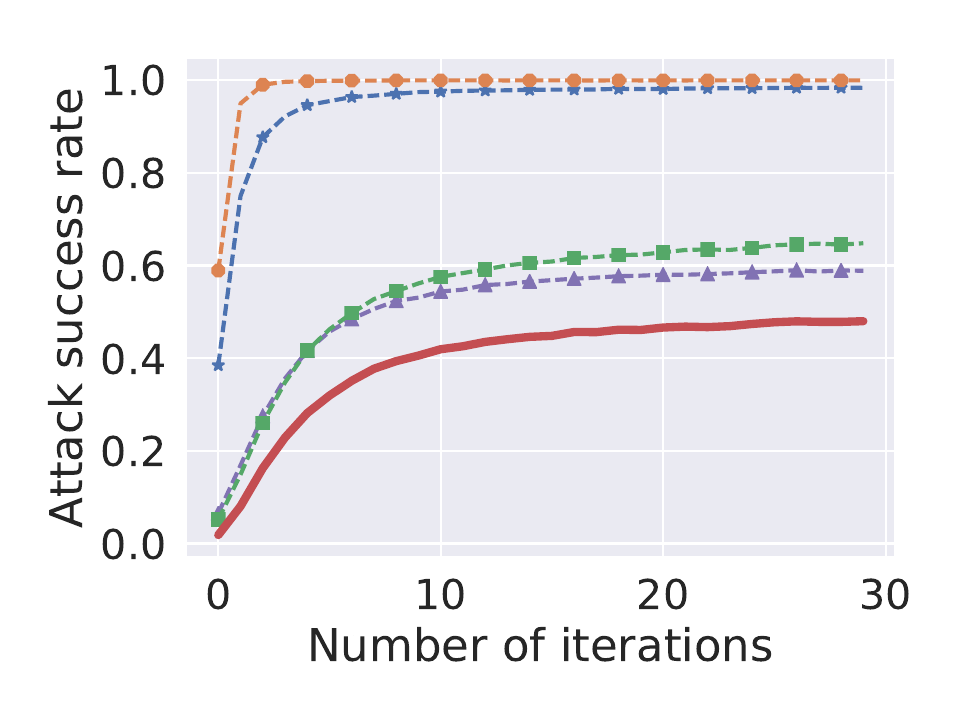} \hspace*{-0.20in} & 
    \includegraphics[width=0.31\textwidth]{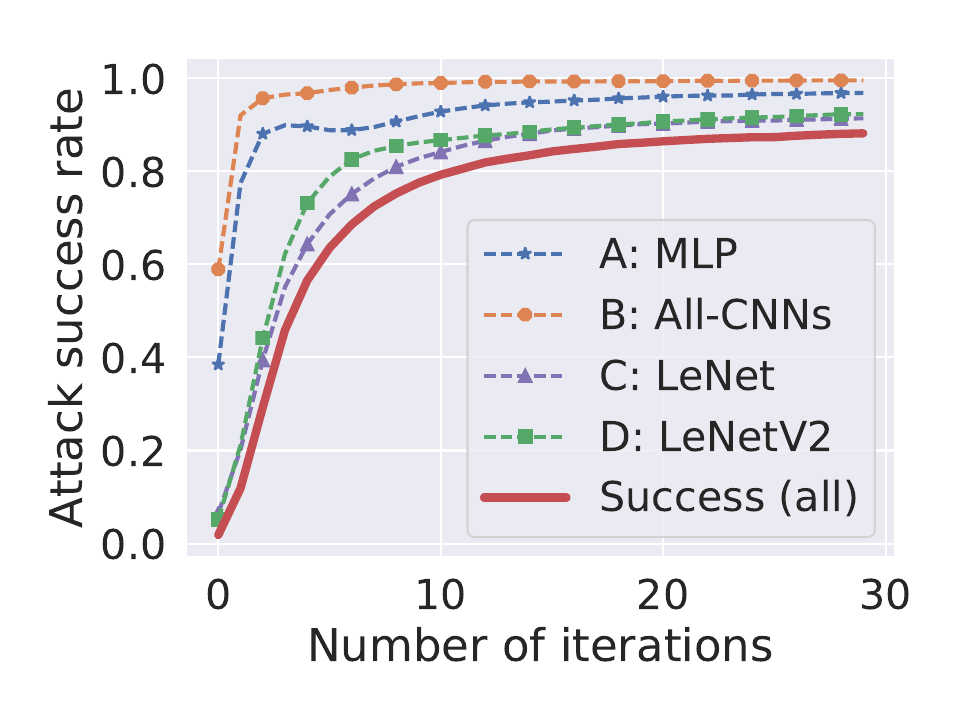} \hspace*{-0.20in} & 
    \includegraphics[width=0.35\textwidth]{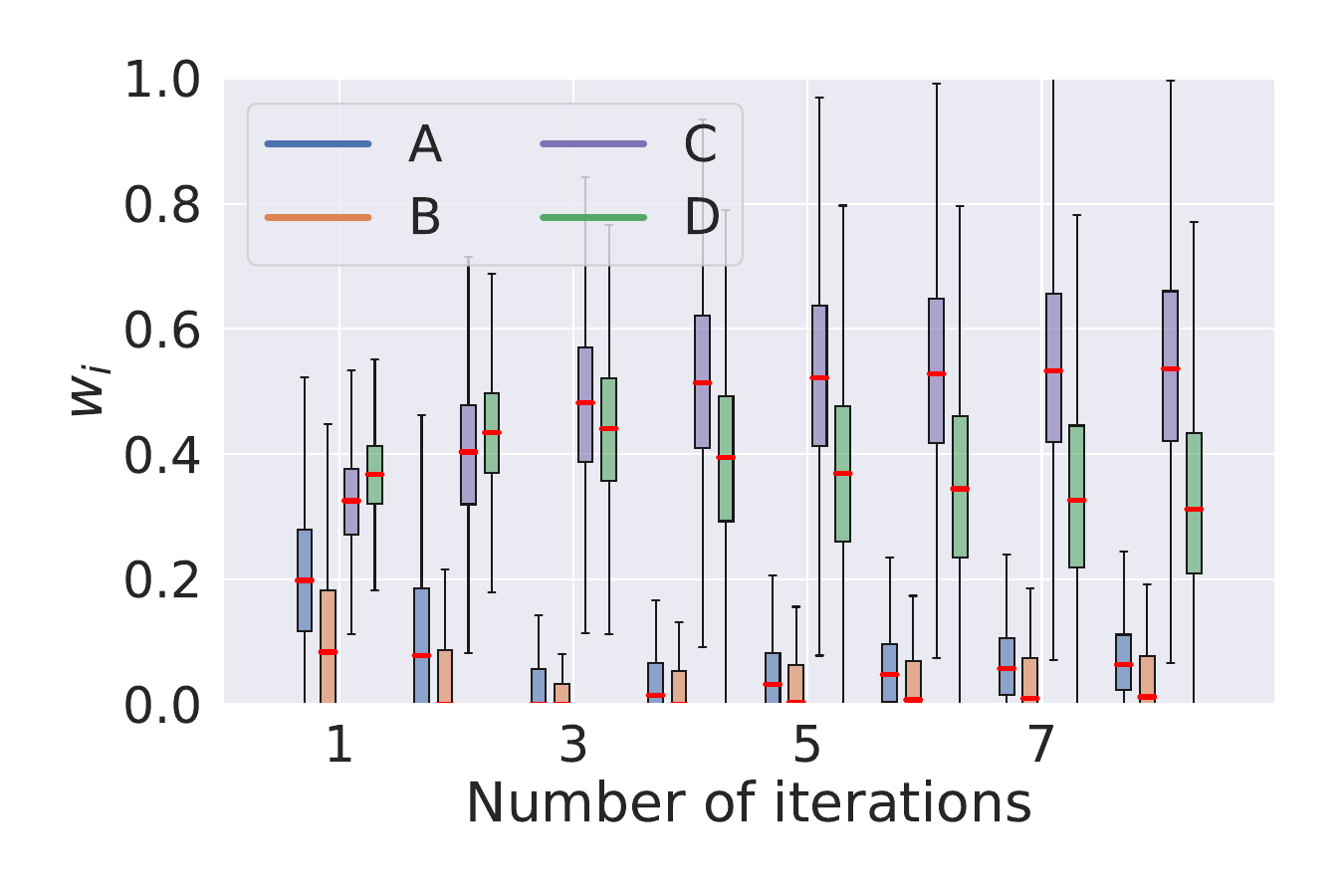} \vspace*{-0.1in}\\
    (a) average case \hspace*{-0.20in} & (b) $\min\max$ \hspace*{-0.20in} & (c) weight $\{w_i\}$ \vspace{-0.1in}
\end{tabular}
\caption{Ensemble attack against four DNN models on MNIST. (a) \& (b): Attack success rate of adversarial examples generated by average PGD or min-max (APGDA) attack method. (c): Boxplot of weight $w$ in min-max adversarial loss. Here we adopt the same $\ell_\infty$-attack as Table~\ref{tab:ensemble_mnist}.}
\label{fig:mnist_ensemble}
\end{figure*}

\subsection{Ensemble Attack over Multiple Models} 

We craft adversarial examples against an ensemble of known classifiers. Recent work~\cite{caad_ensemble} proposed an average ensemble PGD attack, which assumed equal importance among different models, namely, $w_i = 1/K$  in problem \eqref{eq: prob0_ensemble}. Throughout this task, we measure the attack performance via \textbf{ASR}$\boldsymbol{_{all}}$ - the attack success rate (ASR) of fooling model ensembles simultaneously. Compared to the average PGD attack, our approach results in 40.79\% and 17.48\%  ASR$\boldsymbol{_{all}}$ improvement averaged over different $\ell_p$-norm constraints on MNIST and CIFAR-10, respectively. 
In what follows, we provide more detailed results and analysis.

\begin{figure}[t]
\begin{minipage}{\textwidth}
\begin{minipage}[c]{0.49\textwidth}
\begin{table}[H]
\centering
\caption{Comparison of average and min-max (APGDA) ensemble attack on MNIST. %
}
\vspace{-2.5mm}
\vskip 0.05in
\label{tab:ensemble_mnist}
\resizebox{\linewidth}{!}{
\begin{tabular}{@{}c|c|cccc|cccc|c|c@{}}
\toprule
\ \ Box constraint & Opt. & Acc$_A$ & Acc$_B$ & Acc$_C$ & Acc$_D$ & ASR$_{all}$ & Lift ($\uparrow$)\ \\ \midrule
\multirow{2}{*}{$\ell_0$ ($\epsilon=30$)} & $avg.$ & 7.03 & 1.51 & 11.27 & 2.48 & 84.03 & - \\
 & $\min\max$ & 3.65 & 2.36 & 4.99 & 3.11 & \textbf{91.97} & \textbf{9.45\%} \\ \midrule
\multirow{2}{*}{$\ell_1$ ($\epsilon=20$)} & $avg.$ & 20.79 & 0.15 & 21.48 & 6.70 & 69.31 & - \\
 & $\min\max$ & 6.12 & 2.53 & 8.43 & 5.11 & \textbf{89.16} & \textbf{28.64\%} \\ \midrule
\multirow{3}{*}{$\ell_2$ ($\epsilon=3.0$)} & $avg.$ & 6.88 & 0.03 & 26.28 & 14.50 & 69.12 & - \\
 & $\min\max$ & 1.51 & 0.89 & 3.50 & 2.06 & \textbf{95.31} & \textbf{37.89\%} \\ \midrule
\multirow{3}{*}{$\ell_\infty$ ($\epsilon=0.2$)} & $avg.$ & 1.05 & 0.07 & 41.10 & 35.03 & 48.17 & - \\
 & $\min\max$ & 2.47 & 0.37 & 7.39 & 5.81 & \textbf{90.16} & \textbf{87.17\%} \\  \bottomrule
\end{tabular}
}
\vspace{-4mm}
\end{table}
\end{minipage}
\hfill
\begin{minipage}[c]{0.49\textwidth}
\begin{table}[H]
\centering
\caption{Comparison to heuristic weighting schemes on MNIST ($\ell_\infty$-attack, $\epsilon=0.2$). %
}
\vspace{-2.2mm}
\vskip 0.05in
\label{tab:ensemble_heuristic}
\resizebox{\linewidth}{!}{
\begin{tabular}{@{}c|cccc|c|ccc@{}}
\toprule
Opt. & Acc$_A$ & Acc$_B$ & Acc$_C$ & Acc$_D$ & ASR$_{avg}$ & ASR$_{all}$ & Lift ($\uparrow$)\ \\ \midrule
$avg.$ & 1.05 & 0.07 & 41.10 & 35.03 & 80.69 & 48.17 & - \\
 $w_{c+d}$ & 60.37 & 19.55 & 15.10 & 1.87 & 75.78 & 29.32 & -39.13\% \\ 
 $w_{a+c+d}$ & 0.46 & 21.57 & 25.36 & 13.84 & 84.69 & 53.39 & 10.84\% \\ \midrule
 $w_{clip}$~\cite{universal_AT} & 0.66 & 0.03 & 23.43 & 13.23 & 90.66 & 71.54 & 48.52\% \\
 $w_{prior}$ & 1.57 & 0.24 & 17.67 & 13.74 & 91.70 & 74.34 & 54.33\% \\ \midrule
 $w_{static}$ & 10.58 & 0.39 & 9.28 & 10.05 & \underline{92.43} & \underline{77.84} & \underline{61.59\%} \\ 
 $\min\max$ & 2.47 & 0.37 & 7.39 & 5.81 & \textbf{95.99} & \textbf{90.16} & \textbf{87.17\%} \\  \bottomrule
\end{tabular}}
\vspace{-4mm}
\end{table}
\end{minipage}
\end{minipage}
\end{figure}

\begin{figure}[t]
\begin{minipage}{\textwidth}
\begin{minipage}[c]{0.48\textwidth}
\begin{table}[H]
\centering
\caption{Comparison of average and min-max (APGDA) ensemble attack on CIFAR-10. %
}
\label{tab:ensemble_cifar1}
\vspace{-2.5mm}
\vskip 0.05in
\resizebox{\textwidth}{!}{
\begin{tabular}{@{}c|c|cccc|cccc|c|c@{}}
\toprule
Box constraint & Opt. & Acc$_A$ & Acc$_B$ & Acc$_C$ & Acc$_D$ & ASR$_{all}$ & Lift ($\uparrow$)\ \\ \midrule
\multirow{2}{*}{$\ell_0$ ($\epsilon=50$)} & $avg.$ & 27.86 & 3.15 & 5.16 & 6.17 & 65.16 & - \\
 & $\min\max$ & 18.74 & 8.66 & 9.64 & 9.70 & \textbf{71.44} & \textbf{9.64\%} \\ \midrule
\multirow{2}{*}{$\ell_1$ ($\epsilon=30$)} & $avg.$ & 32.92 & 2.07 & 5.55 & 6.36 & 59.74 & - \\
 & $\min\max$ & 12.46 & 3.74 & 5.62 & 5.86 & \textbf{78.65} & \textbf{31.65\%} \\ \midrule
\multirow{2}{*}{$\ell_2$ ($\epsilon=2.0$)} & $avg.$ & 24.3 & 1.51 & 4.59 & 4.20 & 69.55 & - \\
 & $\min\max$ & 7.17 & 3.03 & 4.65 & 5.14 & \textbf{83.95} & \textbf{20.70\%} \\ \midrule
\multirow{2}{*}{$\ell_\infty$ ($\epsilon=0.05$)} & $avg.$ & 19.69 & 1.55 & 5.61 & 4.26 & 73.29 & - \\
 & $\min\max$ & 7.21 & 2.68 & 4.74 & 4.59 & \textbf{84.36} & \textbf{15.10\%} \\  \bottomrule
\end{tabular}}
\vspace{-4mm}
\end{table}
\end{minipage}
\hfill
\begin{minipage}[c]{0.5\textwidth}
\begin{table}[H]
\centering
\caption{Comparison to heuristic weighting schemes on CIFAR-10 ($\ell_\infty$-attack, $\epsilon=0.05$). %
}
\vspace{-2.2mm}
\vskip 0.05in
\label{tab:ensemble_heuristic_cifar}
\resizebox{\linewidth}{!}{
\begin{tabular}{@{}c|c|cccc|ccc@{}}
\toprule
Opt. & Acc$_A$ & Acc$_B$ & Acc$_C$ & Acc$_D$ & ASR$_{avg}$ & ASR$_{all}$ & Lift ($\uparrow$)\ \\ \midrule
 $avg.$ & 19.69 & 1.55 & 5.61 & 4.26 & 92.22 & 73.29 & - \\
 $w_{b+c+d}$ & 42.12 & 1.63 & 5.93 & 4.42 & 75.78 & 51.63 & -29.55\% \\ 
 $w_{a+c+d}$ & 13.33 & 32.41 & 4.83 & 5.44 & 84.69 & 56.89 & -22.38\% \\ \midrule
 $w_{clip}$~\cite{universal_AT} & 11.13 & 3.75 & 6.66 & 6.02 & 90.66 & 77.82 & 6.18\% \\
 $w_{prior}$ & 19.72 & 2.30 & 4.38 & 4.29 & 91.70 & 73.45 & 0.22\% \\ \midrule
 $w_{static}$ & 7.36 & 4.48 & 5.03 & 6.70 & \underline{92.43} & \underline{81.04} & \underline{10.57\%} \\ 
 $\min\max$ & 7.21 & 2.68 & 4.74 & 4.59 & \textbf{95.20} & \textbf{84.36} & \textbf{15.10\%} \\  \bottomrule
\end{tabular}}
\vspace{-4mm}
\end{table}
\end{minipage}
\end{minipage}
\end{figure}

In Table~\ref{tab:ensemble_mnist} and Table~\ref{tab:ensemble_cifar1}, we show that AMGDA significantly outperforms average PGD in ASR$_{all}$.
Taking $\ell_\infty$-attack on MNIST as an example, our min-max attack leads to a 90.16\% ASR$_{all}$, which largely outperforms 48.17\%. The reason is that Model C, D are more difficult to attack, which can be observed from their 
higher test accuracy on adversarial examples. 
As a result, although the adversarial examples crafted by assigning equal weights over multiple models are able to attack \{A, B\} well, they achieve a much lower ASR in \{C, D\}. By contrast, APGDA automatically handles the worst case \{C, D\} by slightly sacrificing the performance on \{A, B\}: 31.47\% averaged ASR improvement on \{C, D\} versus 0.86\% degradation on \{A, B\}.
The choices of $\alpha, \beta, \gamma$ for all experiments and more results on CIFAR-10 are provided in Appendix~\ref{ap:craft_adv_examples} and Appendix~\ref{ap:sec_robust_attacks}.

\begin{wrapfigure}{r}{0.5\textwidth}
  \vspace{-3mm}
  \begin{minipage}{0.25\textwidth}
    \includegraphics[width=\textwidth]{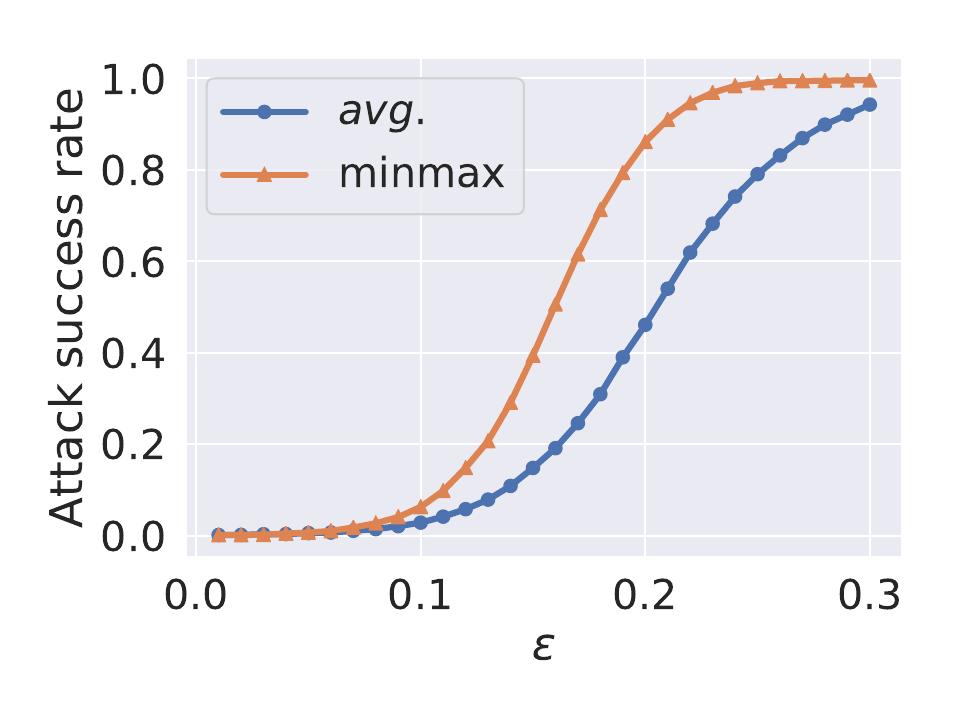}
    \vspace{-6mm}
  \end{minipage}
  \hspace{-4mm}
  \begin{minipage}{0.25\textwidth}
    \includegraphics[width=\textwidth]{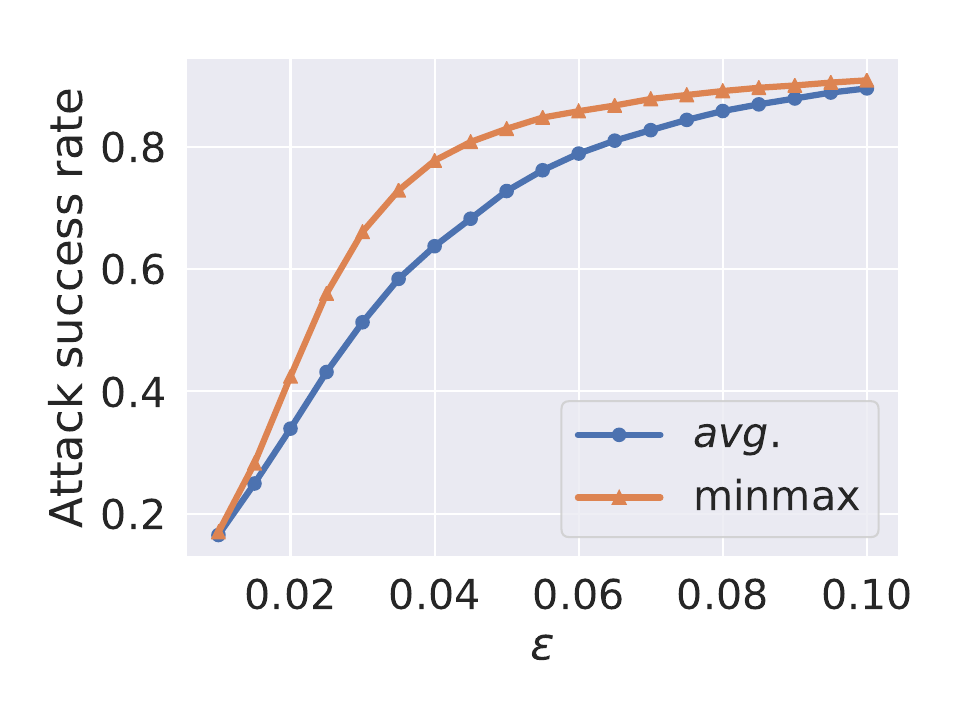}
    \vspace{-6mm}
  \end{minipage}
  \vskip -0.01in
  \caption{ASR of 
  average and min-max $\ell_\infty$ ensemble attack versus maximum perturbation magnitude $\epsilon$. Left (MNIST), Right (CIFAR-10).}
  \label{fig:ensemble_eps_main}
  \vskip -0.13in
\end{wrapfigure}

\paragraph{Effectiveness of learnable domain weights:}
Figure~\ref{fig:mnist_ensemble} depicts the ASR of four models under average/min-max attacks as well as the distribution of domain weights during attack generation. 
For average PGD (Figure \ref{fig:mnist_ensemble}a), Model C and D are attacked insufficiently, leading to relatively low ASR and thus   weak ensemble performance. By contrast, APGDA (Figure \ref{fig:mnist_ensemble}b) will encode the difficulty level to attack different models based on the current attack loss. It dynamically adjusts the weight $w_i$ as shown in Figure~\ref{fig:mnist_ensemble}c. For instance, the weight for Model D is first raised to $0.45$ because D is difficult to attack initially. Then it decreases to $0.3$  once Model D encounters the sufficient attack power and the corresponding attack performance is no longer improved.
It is worth noticing that APGDA is highly efficient because $w_i$ converges after a small number of iterations. 
Figure~\ref{fig:mnist_ensemble}c also shows $w_c > w_d > w_a > w_b$ -- indicating a decrease in model robustness for C, D, A and B, which is exactly verified by Acc$_C$ $>$ Acc$_D$ $>$ Acc$_A$ $>$ Acc$_B$ in the last row of Table~\ref{tab:ensemble_mnist} ($\ell_\infty$-norm).
As the perturbation radius $\epsilon$ varies, we also observe that the ASR of min-max strategy is consistently better or on part with the average strategy (see Figure~\ref{fig:ensemble_eps_main}). 

\begin{table}[t]
\caption{Comparison of average and minmax optimization on universal perturbation over multiple input examples. 
$K$ represents the number of images in each group. ASR$_{avg}$ and ASR$_{all}$ mean attack success rate (\%) of all images and success rate of attacking all the images in each group, respectively.
The adversarial examples are generated by 20-step $\ell_\infty$-APGDA with $\alpha = \frac{1}{6}, \beta = \frac{1}{50}$ and $\gamma = 4$.
}
\vspace{-1mm}
\label{tab:universal_perturbation}
\centering
\resizebox{\textwidth}{!}{%
\setlength\tabcolsep{2.0pt}
\begin{tabular}{@{}ccc|ccc|ccc|ccc|ccc@{}}
\toprule
\multicolumn{3}{c}{Setting} & \multicolumn{3}{c}{$K = 2$} & \multicolumn{3}{c}{$K = 4$} & \multicolumn{3}{c}{$K = 5$} & \multicolumn{3}{c}{$K = 10$} \\ \midrule
\multicolumn{1}{c|}{Dataset} & \multicolumn{1}{c|}{Model} & \multicolumn{1}{c|}{Opt.} & ASR$_{avg}$ & ASR$_{all}$ & Lift ($\uparrow$) & ASR$_{avg}$ & ASR$_{all}$ & Lift ($\uparrow$) & ASR$_{avg}$ & ASR$_{all}$ & Lift ($\uparrow$) & ASR$_{avg}$ & ASR$_{all}$ & Lift ($\uparrow$) \\ \midrule
\multicolumn{1}{c|}{\multirow{8}{*}{CIFAR-10}} & \multicolumn{1}{c|}{\multirow{2}{*}{All-CNNs}} & \multicolumn{1}{c|}{$avg.$} & 91.09 & 83.08 & - & 85.66 & 54.72 & - & 82.76 & 40.20 & - & 71.22 & 4.50 & - \\
\multicolumn{1}{c|}{} & \multicolumn{1}{c|}{} & \multicolumn{1}{c|}{$\min\max$} & 92.22 & \textbf{85.98} & \textbf{3.49\%} & 87.63 & \textbf{65.80} & \textbf{20.25\%} & 85.02 & \textbf{55.74} & \textbf{38.66\%} & 65.64 & \textbf{11.80} & \textbf{162.2\%} \\ \cmidrule(l){2-15} 
\multicolumn{1}{c|}{} & \multicolumn{1}{c|}{\multirow{2}{*}{LeNetV2}} & \multicolumn{1}{c|}{$avg.$} & 93.26 & 86.90 & - & 90.04 & 66.12 & - & 88.28 & 55.00 & - & 72.02 & 8.90 & - \\
\multicolumn{1}{c|}{} & \multicolumn{1}{c|}{} & \multicolumn{1}{c|}{$\min\max$} & 93.34 & \textbf{87.08} & \textbf{0.21\%} & 91.91 & \textbf{71.64} & \textbf{8.35\%} & 91.21 & \textbf{63.55} & \textbf{15.55\%} & 82.85 & \textbf{25.10} & \textbf{182.0\%} \\ \cmidrule(l){2-15} 
\multicolumn{1}{c|}{} & \multicolumn{1}{c|}{\multirow{2}{*}{VGG16}} & \multicolumn{1}{c|}{$avg.$} & 90.76 & 82.56 & - & 89.36 & 63.92 & - & 88.74 & 55.20 & - & 85.86 & 22.40 & - \\
\multicolumn{1}{c|}{} & \multicolumn{1}{c|}{} & \multicolumn{1}{c|}{$\min\max$} & 92.40 & \textbf{85.92} & \textbf{4.07\%} & 90.04 & \textbf{70.40} & \textbf{10.14\%} & 88.97 & \textbf{63.30} & \textbf{14.67\%} & 79.07 & \textbf{30.80} & \textbf{37.50\%} \\ \cmidrule(l){2-15} 
\multicolumn{1}{c|}{} & \multicolumn{1}{c|}{\multirow{2}{*}{GoogLeNet}} & \multicolumn{1}{c|}{$avg.$} & 85.02 & 72.48 & - & 75.20 & 32.68 & - & 71.82 & 19.60 & - & 59.01 & 0.40 & - \\
\multicolumn{1}{c|}{} & \multicolumn{1}{c|}{} & \multicolumn{1}{c|}{$\min\max$} & 87.08 & \textbf{77.82} & \textbf{7.37\%} & 77.05 & \textbf{46.20} & \textbf{41.37\%} & 71.20 & \textbf{33.70} & \textbf{71.94\%} & 45.46 & \textbf{2.40} & \textbf{600.0\%} \\ \bottomrule
\end{tabular}%
}
\end{table}

\paragraph{Comparison with stronger heuristic baselines}

Apart from \emph{average} strategy, 
we compare min-max framework with stronger heuristic weighting scheme in Table~\ref{tab:ensemble_heuristic} (MNIST) and Table~\ref{tab:ensemble_heuristic_cifar} (CIFAR-10).
Specifically, with the prior knowledge of robustness of given models ($C > D > A > B$), we devised several heuristic baselines including: (a) $w_{c+d}$: average PGD on models C and D only; (b) $w_{a+c+d}$: average PGD on models A, C and D only; (c) $w_{clip}$: clipped version of C\&W loss (threshold $\beta = 40$) to balance model weights in optimization as suggested in \cite{universal_AT}; (d) $w_{prior}$: larger weights on the more robust models, $w_{prior} = [w_A, w_B, w_C, w_D] = [0.2, 0.1, 0.4, 0.3]$; (e) $w_{static}$: the converged mean weights of min-max (APGDA) ensemble attack.
For $\ell_2$ ($\epsilon=3.0$) and $\ell_\infty$ ($\epsilon=0.2$) attacks, $w_{static} = [w_A, w_B, w_C, w_D]$ are $[0.209,0.046,0.495,0.250]$ and $[0.080,0.076,0.541,0.303]$, respectively. Table~\ref{tab:ensemble_heuristic} shows that our approach achieve substantial improvement over baselines consistently. Moreover, we highlight that the use of learnable $\mathbf w$ avoids supervised manual adjustment on the heuristic weights or the choice of clipping threshold.  Also, we show that even adopting converged min-max weights statically leads to a huge performance drop on attacking model ensembles, which again verifies the power of dynamically optimizing domain weights during attack generation process. 

\newcolumntype{M}[1]{>{\centering\arraybackslash}m{#1}}
\begin{table}[t]
\caption[Interpretability of domain weight $w$ for universal perturbation (digits 0-4, Sec.~\ref{sec:robust_attack})]{
Interpretability of domain weight $w$ for universal perturbation to multiple inputs on MNIST (\textit{Digit 0, 2, 4}). Domain weight $w$ for different images under $\ell_p$-norm ($p = 0, 1, 2, \infty$). 
}
\vspace{-1mm}
\label{tab:uni_interpret_short}
\centering
\setlength\tabcolsep{1.5pt}
\resizebox{1.0\textwidth}{!}{
\begin{tabular}{@{}c|c|M{1.0cm} M{1.0cm} M{1.0cm} M{1.0cm} M{1.0cm}|
                       M{1.0cm} M{1.0cm} M{1.0cm} M{1.0cm} M{1.0cm}|  M{1.0cm} M{1.0cm} M{1.0cm} M{1.0cm} M{1.0cm} @{}}
\toprule
\multicolumn{2}{c|}{Image} & 
\includegraphics[height=0.39in]{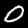} & \includegraphics[height=0.39in]{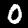} & \includegraphics[height=0.39in]{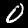} & \includegraphics[height=0.39in]{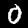} & \includegraphics[height=0.39in]{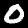} & \includegraphics[height=0.39in]{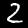} & \includegraphics[height=0.39in]{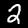} & \includegraphics[height=0.39in]{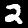} & \includegraphics[height=0.39in]{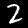} & \includegraphics[height=0.39in]{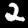} & \includegraphics[height=0.39in]{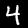} & \includegraphics[height=0.39in]{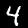} & \includegraphics[height=0.39in]{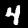} & \includegraphics[height=0.39in]{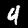} & \includegraphics[height=0.39in]{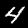} \\ \midrule
\multirow{4}{*}{Weight} & $\ell_0$ & 0. & 0. & 0. & 0. & 1.000 & 0. & 0. & 0.909 & 0. & 0.091 & 0. & 0. & 0.753 & 0. & 0.247 \\
 & $\ell_1$ & 0. & 0. & 0. & 0. & 1.000 & 0. & 0. & 0.843 & 0. & 0.157 & 0.018 & 0. & 0.567 & 0. & 0.416 \\
 & $\ell_2$ & 0. & 0. & 0. & 0. & 1.000 & 0. & 0. & 0.788 & 0. & 0.112 & 0. & 0. & 0.595 & 0. & 0.405 \\
 & $\ell_\infty$ & 0. & 0. & 0. & 0. & 1.000 & 0. & 0. & 0.850 & 0. & 0.150 & 0. & 0. & 0.651 & 0. & 0.349 \\ \midrule
\multirow{2}{*}{Metric} & dist.(C\&W $\ell_2$) & 1.839 & 1.954 & 1.347 & 1.698 & \textbf{3.041} & 1.928 & 1.439 & 2.312 & 1.521 & \textbf{2.356} & 1.558 & 1.229 & \textbf{1.939} & 0.297 & 1.303 \\
 & $\epsilon_{\min}$ ($\ell_\infty$) & 0.113 & 0.167 & 0.073 & 0.121 & \textbf{0.199} & 0.082 & 0.106 & \textbf{0.176} & 0.072 & 0.171 & 0.084 & 0.088 & \textbf{0.122} & 0.060 & 0.094  \\ %
 \bottomrule
\end{tabular}
}
\vspace{-2mm}
\end{table}

\subsection{Multi-Image Universal Perturbation} 

We evaluate APGDA in universal perturbation on MNIST and CIFAR-10, where 10,000 test images are randomly divided into equal-size groups ($K$ images per group) for universal perturbation.
We measure two types of ASR (\%), \textbf{ASR$\boldsymbol{_{avg}}$} and \textbf{ASR$\boldsymbol{_{all}}$}. Here the former represents the ASR averaged over all images in all groups, and the latter signifies the ASR averaged over all groups but a successful attack is counted under a more restricted condition: images within each group must be successfully attacked simultaneously by universal perturbation.
In Table~\ref{tab:universal_perturbation}, we compare  the proposed min-max strategy with  the averaging strategy on the attack performance of generated universal perturbations. APGDA always achieves higher ASR$_{all}$  for different values of $K$. When $K=5$, our approach achieves 42.63\% and 35.21\% improvement over the averaging strategy under MNIST and CIFAR-10.
The universal perturbation generated from APGDA can successfully attack `hard' images (on which  the average-based PGD attack  fails) by self-adjusting domain weights, and thus leads to a higher ASR$_{all}$.

\paragraph{Interpreting ``\textit{image robustness}'' with domain weights $\mathbf w$:}
The min-max universal perturbation also offers interpretability of ``\textit{image robustness}'' by associating domain weights with image visualization. Figure~\ref{tab:uni_interpret_short} %
shows an example in which the large domain weight corresponds to the MNIST letter with clear appearance (e.g., bold letter).
To empirically verify the robustness of image, we report two metrics to measure the difficulty of attacking single image: dist. (C\&W $\ell_2$) denotes the the minimum distortion of successfully attacking images using C\&W ($\ell_2$) attack; $\epsilon_{\min}$ ($\ell_\infty$) denotes the minimum perturbation magnitude for $\ell_\infty$-PGD attack.

\begin{table}[t]
\vspace{-2mm}
\caption{Comparison of average and min-max optimization on robust attack over multiple data transformations on CIFAR-10.
Acc (\%) represents the test accuracy of classifiers on adversarial examples (20-step $\ell_\infty$-APGD ($\epsilon=0.03$) with  $\alpha=\frac{1}{2}, \beta=\frac{1}{100}$ and $\gamma=10$) under different transformations.
}
\label{tab:trans_deter}
\centering
\resizebox{0.7\textwidth}{!}{
\setlength\tabcolsep{2.0pt}
\begin{tabular}{@{}c|c|cccccc|cc@{}}
\toprule
Model & Opt. & Acc$_{ori}$ & Acc$_{flh}$ & Acc$_{flv}$ & Acc$_{bri}$ & Acc$_{gam}$ & Acc$_{crop}$ & ASR$_{all}$ & Lift ($\uparrow$) \\ \midrule
\multirow{2}{*}{A} 
& $avg.$ & 10.80 & 21.93 & 14.75 & 11.52 & 10.66 & 20.03 & 55.88 & - \\
& $\min\max$ & 12.14 & 18.05 & 13.61 & 13.52 & 11.99 & 16.78 & \textbf{60.03} & \textbf{7.43\%} \\
  \midrule
\multirow{2}{*}{B} & $avg.$ & 5.49 & 11.56 & 9.51 & 5.43 & 5.75 & 15.89 & 72.21 & - \\
 & $\min\max$ & 6.22 & 8.61 & 9.74 & 6.35 & 6.42 & 11.99 & \textbf{77.43} & \textbf{7.23\%} \\ \midrule
\multirow{2}{*}{C} & $avg.$ & 7.66 & 21.88 & 15.50 & 8.15 & 7.87 & 15.36 & 56.51 & - \\
 & $\min\max$ & 8.51 & 14.75 & 13.88 & 9.16 & 8.58 & 13.35 & \textbf{63.58} & \textbf{12.51\%} \\ \midrule
\multirow{2}{*}{D} & $avg.$ & 8.00 & 20.47 & 13.46 & 7.73 & 8.52 & 15.90 & 61.13 & - \\
 & $\min\max$ & 9.19 & 13.18 & 12.72 & 8.79 & 9.18 & 13.11 & \textbf{67.49} & \textbf{10.40\%} \\ \bottomrule
\end{tabular}
}
\end{table}

\subsection{Robust Attack over Data Transformations} %
EOT
~\citep{athalye18b} achieves state-of-the-art performance in producing adversarial examples robust to data transformations.
From \eqref{eq: prob0_phy}, we could derive EOT as a special case when the weights satisfy $w_i = 1 / K$ (average case). For each input sample (\textit{ori}), %
we transform the image under a series of functions, e.g., flipping horizontally (\textit{flh}) or vertically (\textit{flv}), adjusting brightness (\textit{bri}), performing gamma correction (\textit{gam}) and cropping (\textit{crop}), and group each image with its transformed variants. Similar to universal perturbation, ASR$_{all}$ is reported to measure the ASR over groups of transformed images (each group is successfully attacked signifies successfully attacking an example under all transformers). %
In Table~\ref{tab:trans_deter}, compared to EOT, our approach leads to 9.39\% averaged lift in ASR$\boldsymbol{_{all}}$ over given models on CIFAR-10 by optimizing the weights for various transformations. 
We leave the the results under randomness (e.g., flipping images randomly \textit{w.p.} 0.8; randomly clipping the images at specific range) in Appendix~\ref{ap:sec_robust_attacks}

\section{Extension: Understanding Defense over Multiple Perturbation Domains} %

In this section,  we show that the min-max principle can also be used to %
gain more insights in generalized adversarial training (AT) from a defender's perspective. 
Different from  promoting robustness of  adversarial examples   against the \textit{worst-case attacking environment} (Sec.\,\ref{sec: min_max_atk}), the generalized AT promotes   model's robustness against the \textit{worst-case defending environment}, given by the existence of multiple $\ell_p$ attacks~\cite{tramer2019adversarial}. 
Our approach obtains better performance than prior works~\cite{tramer2019adversarial,msd} and interpretability by introducing the trainable domain weights. %

\subsection{Adversarial Training under Mixed Types of Adversarial Attacks}

Conventional AT is restricted to a single type of norm-ball constrained adversarial attack \cite{madry2017towards}. For example, AT under $\ell_\infty$ attack yields:
\begin{align}
\displaystyle\minimize_{\boldsymbol{\theta}} ~\mathbb E_{(\mathbf x, \mathbf y) \in \mathcal D}  \maximize_{ \| \boldsymbol{\delta} \|_\infty \leq \epsilon }  ~  f_{\mathrm{tr}}(\boldsymbol{\theta}, \boldsymbol{\delta}; \mathbf  x, y), 
\label{eq: adv_train}
\end{align}
where $\boldsymbol{\theta} \in \mathbb R^n$ denotes  model parameters,   $\boldsymbol{\delta}$ denotes   $\epsilon$-tolerant   $\ell_\infty$ attack, 
 and $f_{\mathrm{tr}}(\boldsymbol{\theta}, \boldsymbol{\delta}; \mathbf  x, y) $ is the  training loss under perturbed examples $\{ (\mathbf x  + \boldsymbol{\delta} , y ) \}$.
However, 
there possibly exist blind attacking spots across multiple types of adversarial attacks so that AT under one  attack would not be strong enough against another  attack \cite{araujo2019robust}.  Thus,
an interesting question is how to generalize AT under multiple types of adversarial attacks~\cite{tramer2019adversarial}. 
One possible way is to use the finite-sum formulation in the inner maximization problem of \eqref{eq: adv_train}, 
namely, $\maximize_{
\{ \boldsymbol{\delta}_i \in \mathcal X_i \} }    \frac{1}{K} \sum_{i=1}^K   f_{\mathrm{tr}}(\boldsymbol{\theta}, \boldsymbol{\delta}_i; \mathbf  x, y)$,
where $\boldsymbol{\delta}_i \in \mathcal X_i$ is the $i$th type of adversarial perturbation defined on $\mathcal X_i$, e.g., different $\ell_p$ attacks.

Since we can   map `attack type' to `domain' considered  in \eqref{eq: prob0}, AT can be generalized  
against the \textit{strongest} adversarial attack across  $K$ attack types in order to avoid blind attacking spots:
\begin{align}\label{eq: adv_train_2type2}
 \begin{array}{l}
\displaystyle\minimize_{\boldsymbol{\theta}} ~\mathbb E_{(\mathbf x, \mathbf y) \in \mathcal D} \maximize_{i \in [K]}
\maximize_{
 \boldsymbol{\delta}_i \in \mathcal X_i  }\, f_{\mathrm{tr}}(\boldsymbol{\theta}, \boldsymbol{\delta}_i; \mathbf x, y).
    \end{array}
\end{align}
In
Lemma\,\ref{eq: lemma_adv_train}, we show that problem \eqref{eq: adv_train_2type2} can be equivalently transformed into the  min-max form.

\begin{mylemma}\label{eq: lemma_adv_train}
Problem \eqref{eq: adv_train_2type2} is equivalent to: 
\begin{align}
\displaystyle\underset{{\boldsymbol{\theta}}}{\mathrm{minimize}} ~\mathbb E_{(\mathbf x, \mathbf y) \in \mathcal D}  ~ \underset{{\mathbf w \in \mathcal P,
\{ \boldsymbol{\delta}_i \in \mathcal X_i \} }}{\mathrm{maximize}}  ~   \sum_{i=1}^K w_i f_{\mathrm{tr}}(\boldsymbol{\theta},\boldsymbol{\delta}_i; \mathbf  x, y)    ,
\label{eq: adv_train_2type2_v3}
\end{align}
where $\mathbf w \in \mathbb R^K$ represent domain weights, and $\mathcal P$ has been defined in \eqref{eq: prob0}.
\end{mylemma}

The proof of Lemma\,\ref{eq: lemma_adv_train} is provided in Appendix\,\ref{app: lemma_min_max_max_ap}.
Similar to \eqref{eq: prob0_reg}, 
a strongly concave regularizer $-\gamma/2 \| \mathbf w - \mathbf 1/K \|_2^2$ can be added into the inner maximization problem  of \eqref{eq: adv_train_2type2_v3}
for  boosting the stability of the learning procedure and  
striking a balance between the max and the average attack performance:%
\vspace{-1mm}
\begin{align}\label{eq: adv_train_ampgd}
\begin{array}{l}
   \displaystyle\minimize_{\boldsymbol{\theta}} ~\mathbb E_{(\mathbf x, \mathbf y) \in \mathcal D} \maximize_{\mathbf w \in \mathcal P,
\{ \boldsymbol{\delta}_i \in \mathcal X_i \} }  ~  \psi(\boldsymbol{\theta}, \mathbf w, \{\boldsymbol{\delta}_i \})\\[1pt]
\psi(\boldsymbol{\theta}, \mathbf w, \{\boldsymbol{\delta}_i \}) \Def  \sum_{i=1}^K w_i f_{\mathrm{tr}}(\boldsymbol{\theta},\boldsymbol{\delta}_i; \mathbf  x, y)  - \frac{\gamma}{2} \| \mathbf w - \mathbf 1/ K\|_2^2
\end{array}
\end{align}
\begin{wrapfigure}{R}{0.47\textwidth}
\vspace{-8mm}
\begin{minipage}{0.47\textwidth}
\begin{algorithm}[H]
\caption{AMPGD to solve problem \eqref{eq: adv_train_ampgd}}
\begin{small}
\begin{algorithmic}[1]
\STATE Input: given $\boldsymbol{\theta}^{(0)}$, $\mathbf w^{(0)}$, $\boldsymbol{\delta}^{(0)}$ and $K > 0$.
\FOR{$t =  1,2,\ldots, T$}
\STATE %
given   $ \mathbf w^{(t-1)}$ and $ \boldsymbol{\delta}^{(t-1)}$, perform SGD  
to update  $\boldsymbol{\theta}^{(t)}$  %
\STATE given  $ \boldsymbol{\theta}^{(t)}$, perform $R$-step PGD to update $\mathbf w^{(t)}$ and $ \boldsymbol{\delta}^{(t)}$ 
\ENDFOR  
\end{algorithmic}\label{alg: min_max_AMPGD_main}
\end{small}
\end{algorithm}
\end{minipage}
\end{wrapfigure}

We propose the \underline{\textbf{a}}lternating  \underline{\textbf{m}}ulti-step \underline{\textbf{p}}rojected \underline{\textbf{g}}radient \underline{\textbf{d}}escent  (AMPGD) method (Algorithm~\ref{alg: min_max_AMPGD_main}) to solve problem (\ref{eq: adv_train_ampgd}). Since AMPGD also follows the min-max principles, we defer more details of this algorithm in Appendix~\ref{ap:sec_ampgd}.  %
We finally remark that our formulation of generalized
AT under multiple perturbations covers prior work~\cite{tramer2019adversarial} as special cases ($\gamma=0$ for max case and $\gamma=\infty$ for average case).

\subsection{Generalized AT vs. Multiple $\ell_p$ Attacks}
\label{sec:adv_train}

\begin{figure}[t]
\begin{minipage}{\textwidth}
\begin{minipage}[c]{0.47\textwidth}
\begin{table}[H]
\resizebox{1.0\linewidth}{!}{
\begin{tabular}{@{}lcccc@{}}
\toprule
        & MAX {[}3{]} & AVG {[}3{]} & MSD {[}2{]} & AMPGD           \\ \midrule
Clean Accuracy & 98.6\%      & 99.1\%      & 98.3\%      & 98.3\%          \\ \midrule
$\ell_\infty$ Attacks~\cite{tramer2019adversarial} ($\epsilon=0.3$) & 51.0\%      & 65.2\%      & 62.7\%      & 76.1\%          \\
$\ell_2$ Attacks~\cite{tramer2019adversarial} ($\epsilon=2.0$) & 61.9\%      & 60.1\%      & 67.9\%      & 70.2\%          \\
$\ell_1$ Attacks~\cite{tramer2019adversarial} ($\epsilon=10$) & 52.6\%      & 39.2\%      & 65.0\%      & 67.2\%         \\
All Attacks~\cite{tramer2019adversarial} & 42.1\%      & 34.9\%      & 58.4\%      & \textbf{64.1\%}  \\ 
\midrule
AA (all attacks)~\cite{autoattack} & 36.9\% & 30.5\% & 55.9\% & 59.3\% \\
AA+ (all attacks)~\cite{autoattack} & 34.3\% & 28.8\% & 54.8\% & \textbf{58.3\%} \\
\bottomrule
\end{tabular}
}
\caption{Adversarial robustness on MNIST.}
\label{tab:mnist_defense}
\end{table}
\end{minipage}
\hfill
\begin{minipage}[c]{0.525\textwidth}
\centering
\begin{figure}[H]
\includegraphics[width=0.33\linewidth]{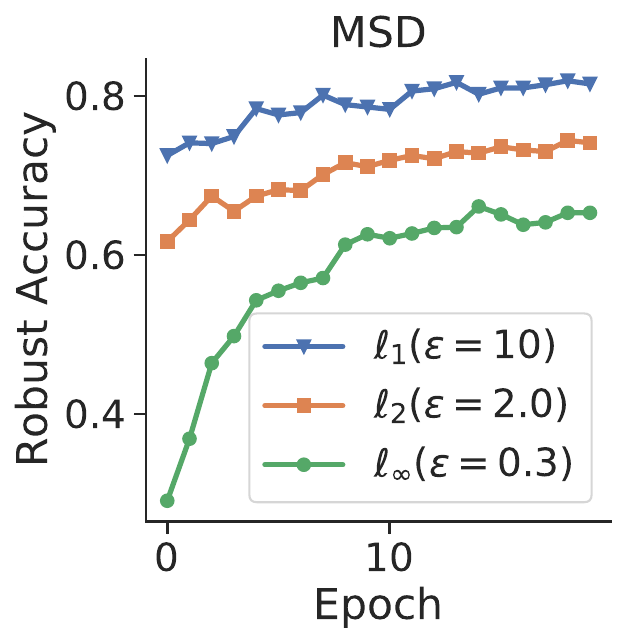}
\hspace{-2mm}
\includegraphics[width=0.33\linewidth]{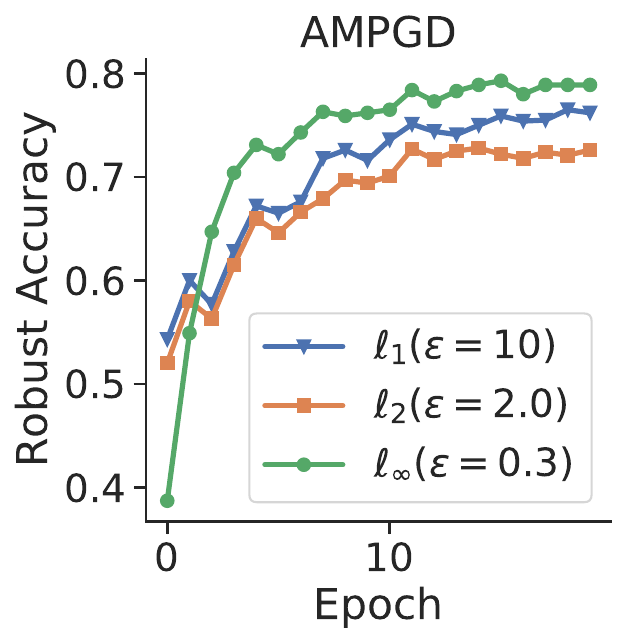}
\hspace{-2mm}
\includegraphics[width=0.33\linewidth]{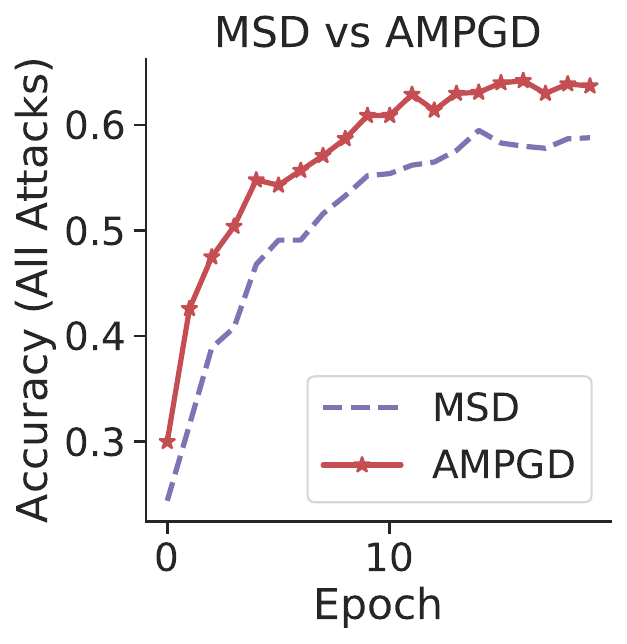}
\vspace{-3mm}
\caption{Robust accuracy of MSD and AMPGD.}
\label{fig:msd_ampgd}
\end{figure}
\end{minipage}
\end{minipage}
\vspace{-5mm}
\end{figure}

\begin{figure}[t]
\begin{minipage}{\textwidth}
\begin{minipage}[c]{0.7\textwidth}
\begin{table}[H]
\centering
\setlength\tabcolsep{2.3pt}
\resizebox{1.00\linewidth}{!}{
\begin{tabular}{@{}lccccccc@{}}
\toprule
  &  $L_\infty$-AT & $L_2$-AT & $L_1$-AT  & MAX~\cite{tramer2019adversarial} & AVG~\cite{tram2018ensemble} & MSD~\cite{msd} & AMPGD           \\ \midrule
Clean Accuracy & 83.3\% & 90.2\% & 73.3\% & 81.0\% & 84.6\% & 81.1\% & 81.5\%         \\ \midrule
$\ell_\infty$ Attacks ($\epsilon=0.03$)~\cite{msd} & 50.7\% & 28.3\% & 0.2\% & 44.9\% & 42.5\% & 48.0\%  & 49.2\%        \\
$\ell_2$ Attacks ($\epsilon=0.5$)~\cite{msd} & 57.3\% & 61.6\% & 0.0\% & 61.7\% & 65.0\% & 64.3\%  & 68.0\%      \\
$\ell_1$ Attacks ($\epsilon=12$)~\cite{msd} &  16.0\% & 46.6\% & 7.9\% & 39.4\% & 54.0\% & 53.0\%  & 50.0\%       \\
All Attacks~\cite{msd} & 15.6\% & 27.5\% & 0.0\% & 34.9\% & 40.6\% & 47.0\% & \textbf{48.7\%} \\  \midrule
AA ($\ell_\infty,\epsilon=0.03$)~\cite{autoattack} & 47.8\% & 22.7\% & 0.0\% & 39.2\% & 40.7\% & 44.4\% & 46.9\% \\ 
AA ($\ell_2,\epsilon=0.5$)~\cite{autoattack} & 57.5\% & 63.1\% & 0.1\% & 62.0\% & 65.5\% & 64.9\% & 64.4\% \\ 
AA ($\ell_1,\epsilon=12$)~\cite{autoattack} & 13.7\% & 23.6\% & 1.4\% & 36.0\% & 58.8\% & 52.4\% & 52.3\% \\
AA (all attacks)~\cite{autoattack} & 12.8\% & 18.4\% & 0.0\%  & 30.8\% & 40.4\% & 44.1\% & \textbf{46.2\%} \\ %
\bottomrule
\end{tabular}
}
\caption{Summary of adversarial accuracy results for CIFAR-10.}
\label{tab:cifar10_defense}
\end{table}
\end{minipage}
\hfill
\begin{minipage}[c]{0.28\textwidth}
\centering
\begin{figure}[H]
\vspace{-1mm}
\includegraphics[width=\linewidth]{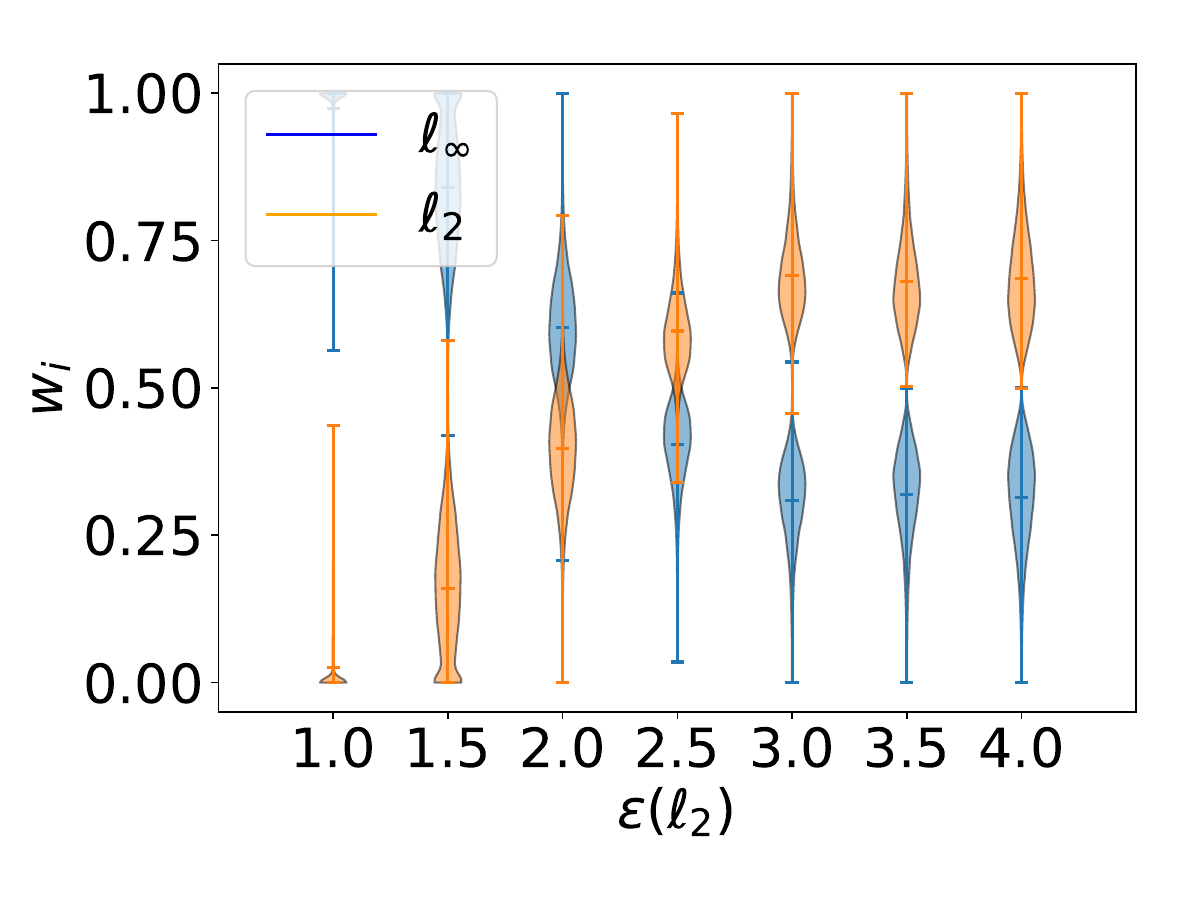}
\vspace{-8mm}
\caption{Domain weights.}
\label{fig:ampgd_weights}
\end{figure}
\end{minipage}
\end{minipage}
\vspace{-3mm}
\end{figure}

Compared to vanilla AT, we show the generalized AT scheme produces model robust to multiple types of perturbation, thus leads to stronger ``overall robustness''.
We present experimental results of generalized AT following~\cite{msd} to achieve simultaneous
robustness to $\ell_\infty$, $\ell_2$, and $\ell_1$ perturbations on the MNIST
and CIFAR-10 datasets.
To the best of our knowledge, MSD proposed in \cite{msd} is the state-of-the-art defense against multiple types of $\ell_p$ attacks.
Specifically, we adopted the same architectures as~\cite{msd} four layer convolutional networks on MNIST and the pre-activation version of the ResNet18~\cite{he2016deep}. 
The perturbation radius $\epsilon$ for $(\ell_\infty, \ell_2, \ell_1)$ balls is set as $(0.3, 2.0, 10)$ and $(0.03, 0.5, 12)$ on MNIST and CIFAR-10 following~\cite{msd}.
Apart from the evaluation $\ell_p$ PGD attacks, we also incorporate the state-of-the-art AutoAttack~\cite{autoattack} for a more comprehensive evaluation under mixed $\ell_p$ perturbations. 

The adversarial accuracy results are reported (higher the better).
As shown in Table~\ref{tab:mnist_defense} and~\ref{tab:cifar10_defense}, our approach outperforms the state-of-the-art defense MSD consistently (4$\sim$6\% and 2\% improvements on MNIST and CIFAR-10).
Compared to MSD that deploys an approximate arg max operation to select the steepest-descent (worst-case) universal perturbation, we leverage the domain weights to self-adjust the strengthens of diverse $\ell_p$ attacks. Thus, we believe that this helps gain supplementary robustness from individual attacks.

\paragraph{Effectiveness of Domain Weights:} 
Figure~\ref{fig:msd_ampgd} shows the robust accuracy curves of MSD and AMPGD on MNIST. As we can see, the proposed AMPGD can quickly adjust the defense strengths to focus on more difficult adversaries - the gap of robust accuracy between three attacks is much smaller. Therefore, it achieves better results by avoiding the trade-off that biases one particular perturbation model at the cost of the others.
In Figure~\ref{fig:ampgd_weights}, we offer deeper insights on how the domain weights work as the strengths of adversary vary. %
Specifically, we consider two perturbation models on MNIST: $\ell_2$ and $\ell_\infty$.
During the training, we fix the $\epsilon$ for $\ell_\infty$ attack during training as 0.2, and change the $\epsilon$ for $\ell_2$ from 1.0 to 4.0. %
As shown in Figure~\ref{fig:ampgd_weights}, the domain weight $w$ increases when the $\ell_2$-attack becomes stronger i.e., $\epsilon(\ell_2)$ increases, which is consistent with min-max spirit -- defending the strongest attack.

\subsection{Additional Discussions}
\label{sec:additional_discussion}
\paragraph{More parameters to tune for min-max?} Our min-max approaches (APGDA and AMPGD) introduce two more hyperparameters - $\beta$ and $\gamma$. However, 
our proposal performs
reasonably well by choosing the learning rate $\alpha$ same as standard PGD and using a large range of regularization coefficient
$\gamma \in [0, 10]$; see Fig. A5 in Appendix. For the learning rate
$\beta$ to update domain weights, we found $1/T$ is usually a very
good practice, where $T$ is the total number of attack iterations.

\paragraph{Time complexity of inner maximization?} Our proposal achieves significant improvements at a low cost of extra computation. Specifically, (1)
our APGDA attack is $1.31\times$ slower than the average PGD; (2) our AMPGD defense is $1.15\times$ slower than average or max AT~\cite{tramer2019adversarial}.

\paragraph{How efficient is the APGDA (Algorithm~\ref{alg: min_max_general}) for solving problem~\eqref{eq: prob0_reg}?}
We remark that the min-max attack generation setup obeys the nonconvex + strongly concave optimization form. Our proposed APGDA is a single-loop algorithm, which is known to achieve a nearly optimal convergence rate for nonconvex-strongly concave min-max optimization~\cite[Table 1]{LinJJ20}.
Furthermore, as our solution gives a natural extension from the commonly-used PGD attack algorithm by incorporating the inner maximization step ~\eqref{eq: pgd_in_max}, it is easy to implement based on existing frameworks.

\paragraph{Clarification on contributions:} 
Our contribution is not to propose a new or more efficient optimization approach for solving min-max optimization problems. Instead, we focus on introducing this formulation to the attack design domain, which has not been studied systematically before. We believe this work is the first solid step to explore the power of min-max principle in the attack design and achieve superior performance on multiple attack tasks.

\section{Conclusion}
In this paper, we revisit the strength of min-max optimization in the context of  adversarial attack generation. %
Beyond adversarial training (AT), we show that many attack generation problems can be  re-formulated in our unified min-max framework, where the maximization is taken over the probability simplex of the set of domains.
Experiments show our min-max attack leads to significant improvements on three tasks. %
Importantly, we demonstrate the self-adjusted domain weights not only stabilize the training procedure but also provides a holistic tool to interpret the risk of different domain sources.
Our min-max principle also helps understand the generalized AT against multiple adversarial attacks. Our approach results in superior performance as well as intepretability.

\section*{Broader Impacts}
Our work provides  a unified framework in design of adversarial examples and robust defenses. The generated adversarial examples can be used to evaluate the robustness of state-of-the-art deep learning vision systems. In spite of different kinds of adversaries, the proposed defense solves one for all by taking into account adversaries' diversity. 
Our work is a beneficial supplement to building trustworthy AI systems, in particular for safety-critical AI applications, such as autonomous vehicles and camera surveillance.
We do not see negative impacts of our work on its ethical aspects and future societal consequences. 

\section*{Acknowledgement}
We sincerely thank the anonymous reviewers for their insightful suggestions and feedback. This work is partially supported by the NSF grant No.1910100, NSF CNS 20-46726 CAR, NSF CAREER CMMI-1750531, NSF ECCS-1609916, and the Amazon Research Award. Resources used in preparing this research were provided, in part, by the Province of Ontario, the Government of Canada through CIFAR, and companies sponsoring the Vector Institute.

\bibliography{reference}
\bibliographystyle{abbrv}

\newpage
\clearpage

\input{supp_neurips21}

\end{document}

%% file: math_commands.tex
\usepackage{amsmath,amsfonts,bm}

\def\eqref#1{(\ref{#1})}

\def\1{\bm{1}}

\DeclareMathAlphabet{\mathsfit}{\encodingdefault}{\sfdefault}{m}{sl}
\SetMathAlphabet{\mathsfit}{bold}{\encodingdefault}{\sfdefault}{bx}{n}

\DeclareMathOperator*{\argmax}{arg\,max}
\DeclareMathOperator*{\argmin}{arg\,min}

%% file: supp_neurips21.tex
\newpage 
\setcounter{section}{0}

\setcounter{section}{0}
\setcounter{figure}{0}
\makeatletter 
\renewcommand{\thefigure}{A\@arabic\c@figure}
\makeatother
\setcounter{table}{0}
\renewcommand{\thetable}{A\arabic{table}}
\setcounter{mylemma}{0}

\appendix

\begin{figure*}
\noindent\rule[0.25in]{\textwidth}{4pt}
\begin{center}
\vskip -0.15in
\Large{\bf 
Supplementary Material\\
\large{Adversarial Attack Generation Empowered by Min-Max Optimization} }  
\end{center}
\vskip 0.15in
\noindent\rule[0.09 in]{\textwidth}{1pt}
\end{figure*}

\begin{abstract}
In this supplementary material, we first provide technical proofs of Proposition~\ref{lemma_Euclidean_projection} and  Lemma~\ref{lemma_Euclidean_projection} in Sec~\ref{app: lemma_Euclidean_projection_ap} and~\ref{app: lemma_min_max_max_ap}. We then discuss the proposed AMPGD algorithm in Sec~\ref{ap:sec_ampgd}. In the next section, we show the details of experimental setup including the model architectures and training details in Sec~\ref{ap:model_arch}, the hyperparameters to craft the adversarial examples (Sec~\ref{ap:craft_adv_examples}), the details of data transformations (Sec~\ref{ap:sec_data_transformation}). Then we show additional experiments results for robust adversarial attacks (Sec~\ref{ap:sec_robust_attacks}) and generalized adversarial training (Sec~\ref{sec:exp_adv_sub}). Finally, we provide more visualizations to show that domain weights $\mathbf w$ provide a holistic tool to interpret ``image robustness'' in Sec~\ref{ap:sec_interpret_w_uni}. The summary of contents in the supplementary is provided in the following. 
\end{abstract}

\etocdepthtag.toc{mtappendix}
\etocsettagdepth{mtchapter}{none}
\etocsettagdepth{mtappendix}{subsection}
\tableofcontents

\newpage
\section{Proof of Proposition~\ref{lemma_Euclidean_projection}}
\label{app: lemma_Euclidean_projection_ap}

\setcounter{myprop}{0}
\begin{myprop}
\label{lemma_Euclidean_projection_full}
Given a point $\mathbf  a \in \mathbb R^d$ and a constraint set $\mathcal X = \{  \boldsymbol{\delta} | \| \boldsymbol{\delta} \|_p \leq \epsilon, \check {\mathbf c} \leq \boldsymbol{\delta} \leq \hat {\mathbf c}\}$, the Euclidean projection $\boldsymbol{\delta}^* = \mathrm{proj}_{\mathcal X} (\mathbf a) $  has the closed-form solution when $p \in \{ 0,1,2 \}$.

1) If $p = 1$, then $\boldsymbol{\delta}^*$ is  given by
\begin{align} \label{eq:  proj_l1}
  {\delta}^*_i =  \left \{ 
    \begin{array}{ll}
        P_{[ \check {c}_i, \hat {c}_i]}( a_i)  &   \sum_{i=1}^{\textcolor{black}{d}} | P_{[ \check {c}_i, \hat {c}_i]}(a_i) | \le   \epsilon  \\
          P_{[ \check {c}_i, \hat {c}_i]}(\mathrm{sign}(a_i)\max{ \{|a_i|-\lambda_1, 0 \}}) &   \text{otherwise},
    \end{array}
    \right. 
\end{align}
where  $\mathbf x_i$ denotes the $i$th element of a vector $\mathbf x$;
$P_{[\check {c}_i, \hat {c}_i]}(\cdot)$ denotes the clip function over the interval $[ \check {c}_i, \hat {c}_i]$; $\mathrm{sign}(x) = 1$ if $x \geq 0$, otherwise $0$; 
$\lambda_1 \in (0, \max_i | a_i|-\epsilon/d]$ is the root of 
$
\sum_{i=1}^d |P_{[ \check {c}_i, \hat {c}_i]}(\mathrm{sign}(a_i)\max{ \{|a_i|-\lambda_1, 0 \}})| = \epsilon
$.

2)  If $p = 2$, then $\boldsymbol{\delta}^*$ is  given by 
\begin{align}\label{eq:  proj_l2}
  {\delta}^*_i =  \left \{ 
    \begin{array}{ll}
        P_{[ \check {c}_i, \hat {c}_i]}( a_i)  &   \sum_{i=1}^d ( P_{[ \check {c}_i, \hat {c}_i]}(a_i) )^2 \le   \epsilon^2  \\
          P_{[ \check {c}_i, \hat {c}_i]}\left ({a_i }/{(\lambda_2+1)} \right ) &   \text{otherwise}, 
    \end{array}
    \right.
\end{align}
where  $\lambda_2 \in (0, \| \mathbf a \|_2/\epsilon-1]$ is the root of 
$
\sum_{i=1}^d   (P_{[ \check {c}_i, \hat {c}_i]} ({ a_i}/{(\lambda_2+1)})  )^2 = \epsilon^2$.

3)  If $p = 0$ and $\epsilon \in \mathbb N_+$, then $\boldsymbol{\delta}^*$ is  given by 
\begin{align}\label{eq:  proj_l0_2}
  {\delta}^*_i =  \left \{ 
    \begin{array}{ll}
        {\delta_i^\prime} &   \eta_i    \ge  {[\boldsymbol{\eta}]_{\epsilon}} \\
        0 &  \text{otherwise},
    \end{array}
    \right.
    \quad  \eta_i =  \left \{ 
        \begin{array}{ll}
         \sqrt{2a_i  {\check c}_i  -  \check {c}_i^2} &   a_i <    \check {c}_i \\ 
            \sqrt{2a_i  {\hat c}_i  -  \hat {c}_i^2} &   a_i >  \hat {c}_i \\ 
          | a_i | & \text{otherwise}.
    \end{array}
    \right.
\end{align}
where $[\boldsymbol{\eta}]_{\epsilon}$ denotes the $\epsilon$-th largest element of $\boldsymbol{\eta}$, and
${\delta}_i^\prime =  P_{[ \check {c}_i, \hat {c}_i]}(a_i)$. 

\end{myprop}

\textbf{\textit{Proof of Proposition~\ref{lemma_Euclidean_projection_full}:}}

\paragraph{$\ell_1$ norm} 
When we find the Euclidean projection of $\mathbf a$ onto the set $\mathcal X$, we solve
\begin{align}\label{eq: L1_ap}
    \begin{array}{ll}
\displaystyle \minimize_{\boldsymbol{\delta}}         & 
\frac{1}{2} \|\boldsymbol{\delta} - \mathbf a \|_2^2 + I_{[\check {\mathbf c},\hat {\mathbf c}]}(\boldsymbol{\delta})  \\
        \st  &  \| \boldsymbol{\delta} \|_1 \le \epsilon,
    \end{array}
\end{align}
where $I_{[\check {\mathbf c},\hat {\mathbf c}]}(\cdot)$ is the indicator function of the set $[\check {\mathbf c},\hat {\mathbf c}]$. The Langragian of this problem is 
\begin{align}
    L &= \frac{1}{2} \|\boldsymbol{\delta} - \mathbf a \|_2^2 + I_{[\check {\mathbf c},\hat {\mathbf c}]}(\boldsymbol{\delta}) + \lambda_1 (\| \boldsymbol{\delta} \|_1 - \epsilon) \\
    &= \sum_{i=1}^d (\frac{1}{2} (\delta_i -  a_i )^2 + \lambda_1 | \delta_i | + I_{[ \check {c}_i, \hat {c}_i]}(\delta_i)) - \lambda_1\epsilon.
\end{align}
The minimizer $\boldsymbol{{\delta}^*}$ minimizes the Lagrangian, it is obtained by elementwise soft-thresholding
\[
 {\delta}^*_i = P_{[\check {c}_i,\hat {c}_i]}(\mathrm{sign}(a_i)\max{ \{|a_i|-\lambda_1, 0 \}}).
\]
where  $\mathbf x_i$ is the $i$th element of a vector $\mathbf x$,
$P_{[\check {c}_i, \hat {c}_i]}(\cdot)$ is the clip function over the interval $[ \check {c}_i, \hat {c}_i]$.

The primal, dual feasibility and complementary slackness are
\begin{align}
    &\lambda_1 = 0, \| \boldsymbol{\delta} \|_1 = \sum_{i=1}^d |{\delta}_i| = \sum_{i=1}^d |P_{[ \check {c}_i, \hat {c}_i]}( a_i)| \le \epsilon  \\
    \bf{or} \ &\lambda_1 > 0, \| \boldsymbol{\delta} \|_1 = \sum_{i=1}^d |{\delta}_i| = \sum_{i=1}^d |P_{[ \check {c}_i, \hat {c}_i]}(\mathrm{sign}( a_i)\max{ \{|a_i|-\lambda_1, 0 \}})| = \epsilon.
\end{align}

If $\sum_{i=1}^{d} | P_{[ \check {c}_i, \hat {c}_i]}(a_i) | \le   \epsilon$, ${\delta}^*_i=P_{[ \check {c}_i, \hat {c}_i]}( a_i)$. Otherwise ${\delta}^*_i = P_{[\check {c}_i,\hat {c}_i]}(\mathrm{sign}(a_i)\max{ \{|a_i|-\lambda_1, 0 \}})$, where $\lambda_1$ is given by the root of the equation
$
\sum_{i=1}^d |P_{[ \check {c}_i, \hat {c}_i]}(\mathrm{sign}(a_i)\max{ \{|a_i|-\lambda_1, 0 \}})| = \epsilon
$. 
Bisection method can be used to solve the above equation for $\lambda_1$, starting with the initial interval $(0, \max_i |a_i|-\epsilon/d]$. Since $\sum_{i=1}^d |P_{[ \check {c}_i, \hat {c}_i]}(\mathrm{sign}(a_i)\max{ \{|a_i| - 0, 0 \}})| = \sum_{i=1}^d | P_{[ \check {c}_i, \hat {c}_i]}(a_i) | > \epsilon$ in this case, and $\sum_{i=1}^d |P_{[ \check {c}_i,\hat {c}_i]}(\mathrm{sign}(a_i)\max{ \{|a_i|- \max_i  |a_i| + \epsilon/d, 0 \}})| \le \sum_{i=1}^d |P_{[\check {c}_i,\hat {c}_i]}(\mathrm{sign}(a_i)(\epsilon/d))| \le \sum_{i=1}^d (\epsilon/d) = \epsilon$. 

\paragraph{$\ell_2$ norm}
 When we find the Euclidean projection of $\mathbf a$ onto the set $\mathcal X$, we solve
\begin{align}\label{eq: L2_ap}
    \begin{array}{ll}
\displaystyle \minimize_{\boldsymbol{\delta}}         & 
\|\boldsymbol{\delta} - \mathbf a \|_2^2 + I_{[\check {\mathbf c},\hat {\mathbf c}]}(\boldsymbol{\delta})  \\
        \st  &  \| \boldsymbol{\delta} \|_2^2 \le \epsilon^2,
    \end{array}
\end{align}
where $I_{[\check {\mathbf c},\hat {\mathbf c}]}(\cdot)$ is the indicator function of the set $[\check {\mathbf c},\hat {\mathbf c}]$. The Langragian of this problem is 
\begin{align}
    L &= \|\boldsymbol{\delta} - \mathbf a \|_2^2 + I_{[\check {\mathbf c}, \hat {\mathbf c}]}(\boldsymbol{\delta}) + \lambda_2 (\| \boldsymbol{\delta} \|_2^2 - \epsilon^2) \\
    &= \sum_{i=1}^d (({\delta_i} -  a_i )^2 + \lambda_2  {\delta_i^2}   + I_{[ \check {c}_i, \hat {c}_i]}({\delta_i})) - \lambda_2\epsilon^2.
\end{align}
The minimizer $\boldsymbol{{\delta}^*}$ minimizes the Lagrangian, it is
\[
 {\delta}^*_i = P_{[\check {c}_i, \hat {c}_i]}(\frac{1}{\lambda_2+1} a_i).
\]
The primal, dual feasibility and complementary slackness are
\begin{align}
    &\lambda_2 = 0, \| \boldsymbol{\delta} \|_2^2 = \sum_{i=1}^d {\delta}_i^2 = \sum_{i=1}^d (P_{[\check {c}_i,\hat {c}_i]}(a_i))^2 \le \epsilon^2  \\
    \bf{or} \ &\lambda_2 > 0, \| \boldsymbol{\delta} \|_2^2 = \sum_{i=1}^d {\delta}_i^2 =  (P_{[\check {c}_i,\hat {c}_i]}(\frac{1}{\lambda_2+1} a_i))^2 = \epsilon^2.
\end{align}

If $\sum_{i=1}^d ( P_{[ \check {c}_i, \hat {c}_i]}(a_i) )^2 \le   \epsilon^2$, ${\delta}^*_i=P_{[ \check {c}_i, \hat {c}_i]}( a_i)$. Otherwise ${\delta}^*_i = P_{[ \check {c}_i, \hat {c}_i]}\left (\frac{1}{\lambda_2+1} a_i \right )$, where $\lambda_2$ is given by the root of the equation
$
\sum_{i=1}^d (P_{[ \check {c}_i, \hat {c}_i]}(\frac{1}{\lambda_2+1} a_i))^2 = \epsilon^2
$. 
Bisection method can be used to solve the above equation for $\lambda_2$, starting with the initial interval $(0, \sqrt{\sum_{i=1}^d (a_i)^2}/\epsilon-1]$. Since $\sum_{i=1}^d (P_{[\check {c}_i,\hat {c}_i]}(\frac{1}{0+1}  a_i))^2 = \sum_{i=1}^d (P_{[ \check {c}_i, \hat {c}_i]}( a_i))^2 > \epsilon^2$ in this case, and $\sum_{i=1}^d (P_{[\check {c}_i,\hat {c}_i]}(\frac{1}{\lambda_2+1} a_i))^2 = \sum_{i=1}^d (P_{[\check {c}_i,\hat {c}_i]}(\epsilon a_i/\sqrt{\sum_{i=1}^d (a_i)^2} ))^2 \le \epsilon^2\sum_{i=1}^d (a_i)^2/(\sqrt{\sum_{i=1}^d (a_i)^2})^2 = \epsilon^2$.

\paragraph{$\ell_0$ norm} For $\ell_0$ norm in $\mathcal X$, it is independent to the box constraint. So we can clip $\mathbf a$ to the box constraint first, which is ${\delta}_i^\prime =  P_{[ \check {c}_i, \hat {c}_i]}(a_i)$, and then project it onto $\ell_0$ norm. 

We find the additional Euclidean distance of every element in $\mathbf a$ and zero after they are clipped to the box constraint, which is
 \begin{align}\label{eq:  proj__l0_ap}
   \eta_i =  \left \{ 
    \begin{array}{ll}
         \sqrt{a_i^2 - (a_i -  \check {c}_i)^2} &   a_i <    \check {c}_i \\ [0.1cm]
            \sqrt{a_i^2 - (a_i -  \hat {c}_i)^2} &   a_i >  \hat {c}_i \\  [0.1cm]
          | a_i | & \text{otherwise}.
    \end{array}
    \right.
\end{align}
It can be equivalently written as 
 \begin{align}
         \eta_i =  \left \{
        \begin{array}{ll}
         \sqrt{2a_i  {\check c}_i  -  \check {c}_i^2} &   a_i <    \check {c}_i \\ 
            \sqrt{2a_i  {\hat c}_i  -  \hat {c}_i^2} &   a_i >  \hat {c}_i \\ 
          | a_i | & \text{otherwise}.
    \end{array}
              \right.
\end{align}
To derive the Euclidean projection onto $\ell_0$ norm, we find the $\epsilon$-th largest element in $\boldsymbol{\eta}$ and call it ${[\boldsymbol{\eta}]_{\epsilon}}$. We keep the elements whose corresponding $\eta_i$ is above or equals to $\epsilon$-th, and set rest to zeros. The closed-form solution is given by 
 \begin{align}\label{eq:  proj_l0_2_ap}
   {\delta}^*_i =  \left \{ 
    \begin{array}{ll}
        {\delta_i^\prime} &   \eta_i    \ge  {[\boldsymbol{\eta}]_{\epsilon}} \\
        0 &  \text{otherwise}.
    \end{array}
    \right.
\end{align}
\hfill $\square$

\textbf{Difference with \citep[Proposition 4.1]{hein2017formal}.}
We remark that  \cite{{hein2017formal}} discussed a relevant problem of generating $\ell_p$-norm based adversarial examples under  box and linearized classification constraints. The key difference between our proof  and that of \citep[Proposition 4.1]{hein2017formal} 
is summarized below.
First,  we place $\ell_p$ norm as a hard constraint rather than minimizing it in the objective function. This difference will make our Lagrangian function   more involved with a newly introduced non-negative Lagrangian multiplier.  Second, the problem of our interest is projection  onto the intersection of box and $\ell_p$ constraints. Such a projection step can then be  combined with an attack loss (no need of linearization) for generating adversarial examples. Third, we cover the case of $\ell_0$ norm.

\newpage 

\section{Proof of Lemma \ref{eq: lemma_adv_train}}
\label{app: lemma_min_max_max_ap}
\begin{mylemma}\label{eq: lemma_adv_train_ap}
Problem \eqref{eq: adv_train_2type2} is equivalent to 

{
\vspace{-8mm}
\begin{align*}
\begin{array}{l}
\displaystyle\underset{{\boldsymbol{\theta}}}{\mathrm{minimize}} ~\mathbb E_{(\mathbf x, \mathbf y) \in \mathcal D}  ~ \underset{{\mathbf w \in \mathcal P,
\{ \boldsymbol{\delta}_i \in \mathcal X_i \} }}{\mathrm{maximize}}  ~   \sum_{i=1}^K w_i f_{\mathrm{tr}}(\boldsymbol{\theta},\boldsymbol{\delta}_i; \mathbf  x, y)    ,
    \vspace{-2mm}
    \end{array}
\end{align*}
\vspace{-2mm}
}
where $\mathbf w \in \mathbb R^K$ represent domain weights, and $\mathcal P$ has been defined in \eqref{eq: prob0}.
\end{mylemma}
\textbf{\textit{Proof of Lemma~\ref{eq: lemma_adv_train}:}}

Similar to \eqref{eq: prob0},
problem \eqref{eq: adv_train_2type2} is equivalent to 
\begin{align} %
    \begin{array}{l}
\displaystyle\minimize_{\boldsymbol{\theta}} ~\mathbb E_{(\mathbf x, \mathbf y) \in \mathcal D} \maximize_{ \mathbf w \in \mathcal P} \sum_{i=1}^K w_i F_i(\boldsymbol{\theta}).
    \end{array}
\end{align}

Recall that $F_i(\boldsymbol{\theta}) \Def \maximize_{
\boldsymbol{\delta}_i \in \mathcal X_i  }\, f_{\mathrm{tr}}(\boldsymbol{\theta}, \boldsymbol{\delta}_i; \mathbf x, y) $, problem can then be written as
\begin{align}\label{eq: adv_train_2type2_temp1_ap_v2}
    \begin{array}{l}
\displaystyle\minimize_{\boldsymbol{\theta}} ~\mathbb E_{(\mathbf x, \mathbf y) \in \mathcal D} \maximize_{ \mathbf w \in \mathcal P} \sum_{i=1}^K [ w_i \maximize_{
\boldsymbol{\delta}_i \in \mathcal X_i  }\, f_{\mathrm{tr}}(\boldsymbol{\theta}, \boldsymbol{\delta}_i; \mathbf x, y) ].
    \end{array}
\end{align}
According to proof by contradiction, it is clear that problem \eqref{eq: adv_train_2type2_temp1_ap_v2}
 is  equivalent to 
\begin{align}\label{eq: adv_train_2type2_ap}
    \begin{array}{l}
\displaystyle\minimize_{\boldsymbol{\theta}} ~\mathbb E_{(\mathbf x, \mathbf y) \in \mathcal D}   \maximize_{   
\mathbf w \in \mathcal P, \{ \boldsymbol{\delta}_i \in \mathcal X_i \}
}\, \sum_{i=1}^K w_i f_{\mathrm{tr}}(\boldsymbol{\theta}, \boldsymbol{\delta}_i; \mathbf x, y).
    \end{array}
\end{align}
\hfill $\square$

\newpage 
\section{Alternating Multi-step PGD (AMPGD) for Generalized  AT}
\label{ap:sec_ampgd}

In this section, we present the full \underline{\textbf{a}}lternating  \underline{\textbf{m}}ulti-step \underline{\textbf{p}}rojected \underline{\textbf{g}}radient \underline{\textbf{d}}escent  (AMPGD) algorithm to solve the problem \eqref{eq: adv_train_ampgd}, which is repeated as follows 
\begin{align}\label{eq: adv_train_2type2_v3_reg}
    \begin{array}{l}
\displaystyle\minimize_{\boldsymbol{\theta}} ~\mathbb E_{(\mathbf x, \mathbf y) \in \mathcal D}  ~ \maximize_{\mathbf w \in \mathcal P,
\{ \boldsymbol{\delta}_i \in \mathcal X_i \} }  ~  \psi(\boldsymbol{\theta}, \mathbf w, \{\boldsymbol{\delta}_i \}) \\ \psi(\boldsymbol{\theta}, \mathbf w, \{\boldsymbol{\delta}_i \}) \Def  \sum_{i=1}^K w_i f_{\mathrm{tr}}(\boldsymbol{\theta},\boldsymbol{\delta}_i; \mathbf  x, y)  - \frac{\gamma}{2} \| \mathbf w - \mathbf 1/ K\|_2^2 %
\nonumber
    \end{array} 
\end{align}
\begin{algorithm}[H]
\caption{AMPGD to solve problem \eqref{eq: adv_train_ampgd}}
\resizebox{0.95\textwidth}{!}{
\begin{minipage}{\textwidth}
\begin{algorithmic}[1]
\STATE Input: given $\boldsymbol{\theta}^{(0)}$, $\mathbf w^{(0)}$, $\boldsymbol{\delta}^{(0)}$ and $K > 0$.
\FOR{$t =  1,2,\ldots, T$}
\STATE %
given   $ \mathbf w^{(t-1)}$ and $ \boldsymbol{\delta}^{(t-1)}$, perform SGD  
to update  $\boldsymbol{\theta}^{(t)}$  %
\STATE given  $ \boldsymbol{\theta}^{(t)}$, perform $R$-step PGD to update $\mathbf w^{(t)}$ and $ \boldsymbol{\delta}^{(t)}$ 
\ENDFOR  
\end{algorithmic}\label{alg: min_max_AMPGD}
\end{minipage}
}
\end{algorithm}
\vspace{-4mm}
Problem \eqref{eq: adv_train_ampgd} is in 
a more general non-convex non-concave min-max setting, where the inner maximization 
involves both domain weights $\mathbf w$ and adversarial perturbations $\{ \boldsymbol{\delta}_i \}$. 
It was  shown in \citep{nouiehed2019solving} that 
the multi-step PGD is required for inner maximization   in order to approximate the near-optimal solution.  This is also in  the similar spirit of AT \citep{madry2017towards}, which executed multi-step PGD attack during inner maximization. 
We summarize AMPGD in Algorithm\,\ref{alg: min_max_AMPGD}.
At step\,4 of Algorithm\,\ref{alg: min_max_AMPGD}, each PGD step to update $\mathbf w$ and $\boldsymbol{\delta}$ can be decomposed as
\begin{align}
     &\mathbf w^{(t)}_r = \mathrm{proj}_{\mathcal P} \left  (\mathbf w^{(t)}_{r-1} +\beta  \nabla_{\mathbf w} \psi(\boldsymbol{\theta}^{(t)}, 
     \mathbf w^{(t)}_{r-1},  \{\boldsymbol{\delta}_{i,r-1}^{(t)} \} )  \right ), \forall r \in [R], \nonumber \\
     &\boldsymbol{\delta}^{(t)}_{i,r} = \mathrm{proj}_{\mathcal X_i} \left  ( \boldsymbol{\delta}^{(t)}_{i,r-1} +\beta  \nabla_{\boldsymbol{\delta}} \psi(\boldsymbol{\theta}^{(t)}, 
     \mathbf w^{(t)}_{r-1},  \{\boldsymbol{\delta}_{i,r-1}^{(t)} \} )  \right ), \forall r, i \in [R],   [K] \nonumber
\end{align}
where let $ \mathbf w^{(t)}_{1} \Def  \mathbf w^{(t-1)}$ and $\boldsymbol{\delta}_{i,1}^{(t)} \Def  \boldsymbol{\delta}_{i}^{(t-1)}$. Here the superscript $t$ represents the iteration index of AMPGD, and the subscript $r$ denotes
the iteration index of $R$-step PGD. Clearly, the above projection operations can be   derived  for closed-form expressions through \eqref{eq: pgd_in_max} and Lemma\,\ref{lemma_Euclidean_projection}. To the best of our knowledge, it is still an open question  to build   theoretical convergence guarantees for solving  the general non-convex non-concave min-max problem like \eqref{eq: adv_train_ampgd}, except the work \citep{nouiehed2019solving} which proposed $O(1/T)$ convergence rate if the objective function satisfies a strict Polyak-{\L}ojasiewicz condition  \citep{karimi2016linear}.

\newpage 
\section{Experiment Setup}

\label{app:experiment_setup}

\subsection{Model Architectures and Training Details}
\label{ap:model_arch}
For a comprehensive evaluation of proposed algorithms, we adopt a set of diverse DNN models (Model A to H), including multi-layer perceptrons (MLP), All-CNNs~\cite{allcnns2015}, LeNet~\cite{Lecun1998gradient}, LeNetV2\footnote{An enhanced version of original LeNet with more layers and units (see Table~\ref{tab:arch_ap} Model D).}, VGG16~\cite{vgg16}, ResNet50~\cite{he2016deep}, Wide-ResNet~\cite{madry2017towards} and GoogLeNet~\cite{googlenet}. For the last four models, we use the exact same architecture as original papers and evaluate them only on CIFAR-10 dataset. The details for model architectures are provided in Table~\ref{tab:arch_ap}. For compatibility with our framework, we implement and train these models based on the strategies adopted in pytorch-cifar\footnote{\url{https://github.com/kuangliu/pytorch-cifar}} and achieve comparable performance on clean images; see Table~\ref{tab:benign}. To foster reproducibility, all the trained models are publicly accessible in the anonymous link.
Specifically, we trained MNIST classifiers for 50 epochs with Adam and a constant learning rate of 0.001. For CIFAR-10 classifers, the models are trained for 250 epochs with SGD (using 0.8 nesterov momentum, weight decay $5e^{-4}$). The learning rate is reduced at epoch 100 and 175 with a decay rate of 0.1. The initial learning rate is set as 0.01 for models \{A, B, C, D, H\} and 0.1 for \{E, F, G\}. Note that no data augmentation is employed in the training.

\begin{table}[htb]
\renewcommand{\arraystretch}{1.2}
\caption[Neural network architectures used in the paper (Sec.~\ref{sec:experiments})]{Neural network architectures used on the MNIST and CIFAR-10 dataset. Conv: convolutional layer, FC: fully connected layer, Globalpool: global average pooling layer.}
\centering
\resizebox{0.85\textwidth}{!}{
	\begin{tabular}{@{}cccc@{}} \toprule
		\textbf{A} (MLP) & \textbf{B} (All-CNNs~\cite{allcnns2015})          & \textbf{C} (LeNet~\cite{Lecun1998gradient})    & \textbf{D} (LeNetV2)  \\ \midrule
		FC(128) + Relu   & Conv([32, 64], 3, 3) + Relu    & Conv(6, 5, 5) + Relu  & Conv(32, 3, 3) + Relu \\
		FC(128) + Relu   & Conv(128, 3, 3) + Dropout(0.5) & Maxpool(2, 2)         & Maxpool(2, 2)         \\
		FC(64) + Relu    & Conv([128, 128], 3, 3) + Relu  & Conv(16, 5, 5) + Relu & Conv(64, 3, 3) + Relu \\
		FC(10)           & Conv(128, 3, 3) + Dropout(0.5) & Maxpool(2, 2)         & Maxpool(2, 2)         \\
		Softmax          & Conv(128, 3, 3) + Relu         & FC(120) + Relu        & FC(128) + Relu        \\
		                 & Conv(128, 1, 1) + Relu         & FC(84) + Relu         & Dropout(0.25)         \\
		                 & Conv(10, 1, 1) + Globalpool    & FC(10)                & FC(10)                \\
		                 & Softmax                        & Softmax               & Softmax               \\ \midrule
		\textbf{E} (VGG16~\cite{vgg16}) & \textbf{F} (ResNet50~\cite{he2016deep})          & \textbf{G} (Wide-ResNet~\cite{madry2017towards})    & \textbf{H} (GoogLeNet~\cite{googlenet})  \\ \bottomrule
	\end{tabular}
	\label{tab:arch_ap}
}
\end{table}

\begin{table}[htb]
\renewcommand{\arraystretch}{1.1}
\caption[Clean test accuracy of the DNN models (Sec.~\ref{sec:experiments})]{Clean test accuracy of DNN models on MNIST and CIFAR-10. We roughly derive the model robustness by attacking models separately using FGSM~\cite{goodfellow2014explaining}. The adversarial examples are generated by FGSM $\ell_\infty$-attack ($\epsilon=0.2$).}
\label{tab:benign}
\centering
\resizebox{0.85\textwidth}{!}{%
\begin{tabular}{@{}lcc|lcclcc@{}}
\toprule
\multicolumn{3}{c|}{MNIST} & \multicolumn{6}{c}{CIFAR-10} \\ \midrule
\multicolumn{1}{c}{Model} & Acc. & FGSM & \multicolumn{1}{c}{Model} & Acc. & \multicolumn{1}{c|}{FGSM} & \multicolumn{1}{c}{Model} & Acc. & FGSM \\ \midrule
A: MLP & 98.20\% & 18.92\% & A: MLP & 55.36\% & \multicolumn{1}{c|}{11.25\%} & E: VGG16 & 87.57\% & 10.83\% \\
B: All-CNNs & 99.49\% & 50.95\% & B: All-CNNs & 84.18\% & \multicolumn{1}{c|}{9.89\%} & F: ResNet50 & 88.11\% & 10.73\% \\
C: LeNet & 99.25\% & 63.23\% & C: LeNet & 64.95\% & \multicolumn{1}{c|}{14.45\%} & G: Wide-ResNet & 91.67\% & 15.78\% \\
D: LeNetV2 & 99.33\% & 56.36\% & D: LeNetV2 & 74.89\% & \multicolumn{1}{c|}{9.77\%} & H: GoogLeNet & 90.92\% & 9.91\% \\ \bottomrule
\end{tabular}%
}
\end{table}

\subsection{Crafting Adversarial Examples}
\label{ap:craft_adv_examples}
We adopt variant C\&W loss in APGDA/PGD as suggested in ~\cite{madry2017towards,carlini2017towards} with a confidence parameter $\kappa = 50$. Cross-entropy loss is also supported in our implementation. The adversarial examples are generated by 20-step PGD/APGDA unless otherwise stated (e.g., 50 steps for ensemble attacks). Note that proposed algorithms are robust and will not be affected largely by the choices of hyperparameters ($\alpha, \beta, \gamma$). In consequence, we do not finely tune the parameters on the validation set.
Specifically, The learning rates $\alpha, \beta$ and regularization factor $\gamma$ for Table~\ref{tab:ensemble_mnist} are set as - %
$\ell_0: \alpha=1, \beta=\frac{1}{100}, \gamma=7$, $\ell_1: \alpha=\frac{1}{4}, \beta=\frac{1}{100}, \gamma=5$, $\ell_2: \alpha=\frac{1}{10}, \beta=\frac{1}{100}, \gamma=3$; $\ell_\infty: \alpha=\frac{1}{4}, \beta=\frac{1}{50}, \gamma=3$. %
For Table~\ref{tab:ensemble_cifar1}, the hyper-parameters are set as $\ell_0: \alpha=1, \beta=\frac{1}{150}, \gamma=1$, $\ell_1: \alpha=\frac{1}{4}, \beta=\frac{1}{100}, \gamma=5$, $\ell_2: \alpha=\frac{1}{8}, \beta=\frac{1}{100}, \gamma=3$; $\ell_\infty: \alpha=\frac{1}{5}, \beta=\frac{1}{50}, \gamma=6$.

Due to varying model robustness on different datasets, the perturbation magnitudes $\epsilon$ are set separately~\cite{carlini2019}. For universal perturbation experiments, the $\epsilon$ are set as 0.2 (A, B), 0.3 (C) and 0.25 (D) on MNIST; 0.02 (B, H), 0.35 (E) and 0.05 (D) on CIFAR-10. For generalized AT, the models on MNIST are trained following the same rules in last section, except that training epochs are prolonged to 350 and adversarial examples are crafted for assisting the training with a ratio of 0.5. Our experiment setup is based on CleverHans
package\footnote{\url{https://github.com/tensorflow/cleverhans}} and Carlini and Wagner's framework\footnote{\url{https://github.com/carlini/nn_robust_attacks}}.

\subsection{Details of Conducted Data Transformations}
\label{ap:sec_data_transformation}

To demonstrate the effectiveness of APGDA in generating robust adversarial examples against multiple transformations, we adopt a series of common transformations, including a\&b) flipping images horizontally (\textit{flh}) and vertically (\textit{flv}); c) adjusting image brightness (\textit{bri}); d) performing gamma correction (\textit{gam}), e) cropping and re-sizing images (\textit{crop}); f) rotating images (\textit{rot}). 

Moreover, both deterministic and stochastic transformations are considered in our experiments. In particular, Table~\ref{tab:trans_deter} and Table~\ref{tab:trans_deter_cifar} are deterministic settings - \textit{rot}: rotating images 30 degree clockwise; \textit{crop}: cropping images in the center ($0.8 \times 0.8$) and resizing them to $32 \times 32$; \textit{bri}: adjusting the brightness of images with a scale of 0.1; \textit{gam}: performing gamma correction with a value of 1.3.
Differently, in Table~\ref{tab:trans_stoch_cifar}, we introduce randomness for drawing samples from the distribution - \textit{rot}: rotating images randomly from -10 to 10 degree; \textit{crop}: cropping images in the center randomly (from 0.6 to 1.0); other transformations are done with a probability of 0.8. In experiments, we adopt \texttt{tf.image} API~\footnote{\url{https://www.tensorflow.org/api_docs/python/tf/image}} for processing the images.

\newpage
\section{Additional Experiment Results - Robust adversarial attacks}
\label{ap:sec_robust_attacks}

\subsection{Ensemble Attack over Multiple Models}
\label{ap:sec_ensemble}
Table~\ref{tab:ensemble_cifar1} and
\ref{tab:ensemble_cifar2} shows the performance of average (ensemble PGD~\cite{caad_ensemble}) and min-max (APGDA) strategies for attacking model ensembles. Our min-max approach results in 
19.27\% and 
15.69\% averaged improvement on ASR$_{all}$ over models \{A, B, C, D\} and 
\{A, E, F, H\} on CIFAR-10.

\begin{table}[htbp!]
\centering
\caption{Comparison of average and min-max (APGDA) ensemble attack over models \{A, E, F, H\} on CIFAR-10. Acc (\%) represents the test accuracy of classifiers on adversarial examples. The learning rates $\alpha, \beta$ and regularization factor $\gamma$ are set as - $\ell_0: \alpha=1, \beta=\frac{1}{150}, \gamma=1$, $\ell_1: \alpha=\frac{1}{4}, \beta=\frac{1}{100}, \gamma=5$, $\ell_2: \alpha=\frac{1}{8}, \beta=\frac{1}{100}, \gamma=3$; $\ell_\infty: \alpha=\frac{1}{5}, \beta=\frac{1}{50}, \gamma=6$. The attack iteration for APGDA is set as 50. 
}
\label{tab:ensemble_cifar2}
\resizebox{0.7\textwidth}{!}{
\begin{tabular}{@{}c|c|cccc|cccc|c|c@{}}
\toprule
Box constraint & Opt. & Acc$_A$ & Acc$_E$ & Acc$_F$ & Acc$_H$ & ASR$_{all}$ & Lift ($\uparrow$)\ \\ \midrule
\multirow{2}{*}{$\ell_0$ ($\epsilon=70$)} & $avg.$ & 27.38 & 6.33 & 7.18 & 6.99 & 66.56 & - \\
 & $\min\max$ & 19.38 & 8.72 & 9.48 & 8.94 & \textbf{73.83} & \textbf{10.92\%} \\ \midrule
\multirow{2}{*}{$\ell_1$ ($\epsilon=30$)} & $avg.$ & 30.90 & 2.06 & 1.85 & 1.84 & 66.23 & - \\
 & $\min\max$ & 12.56 & 3.21 & 2.70 & 2.72 & \textbf{83.13} & \textbf{25.52\%} \\ \midrule
\multirow{2}{*}{$\ell_2$ ($\epsilon=1.5$)} & $avg.$ & 20.87 & 1.75 & 1.21 & 1.54 & 76.41 & - \\
 & $\min\max$ & 10.26 & 3.15 & 2.24 & 2.37 & \textbf{84.99} & \textbf{11.23\%} \\ \midrule
\multirow{2}{*}{$\ell_\infty$ ($\epsilon=0.03$)} & $avg.$ & 25.75 & 2.59 & 1.66 & 2.27 & 70.54 & - \\
 & $\min\max$ & 13.47 & 3.79 & 3.15 & 3.48 & \textbf{81.17} & \textbf{15.07\%} \\   \bottomrule
\end{tabular}}
\end{table}

\begin{figure}[htbp!]
\centerline{ 
\begin{tabular}{cccc}
\includegraphics[width=0.35\textwidth]{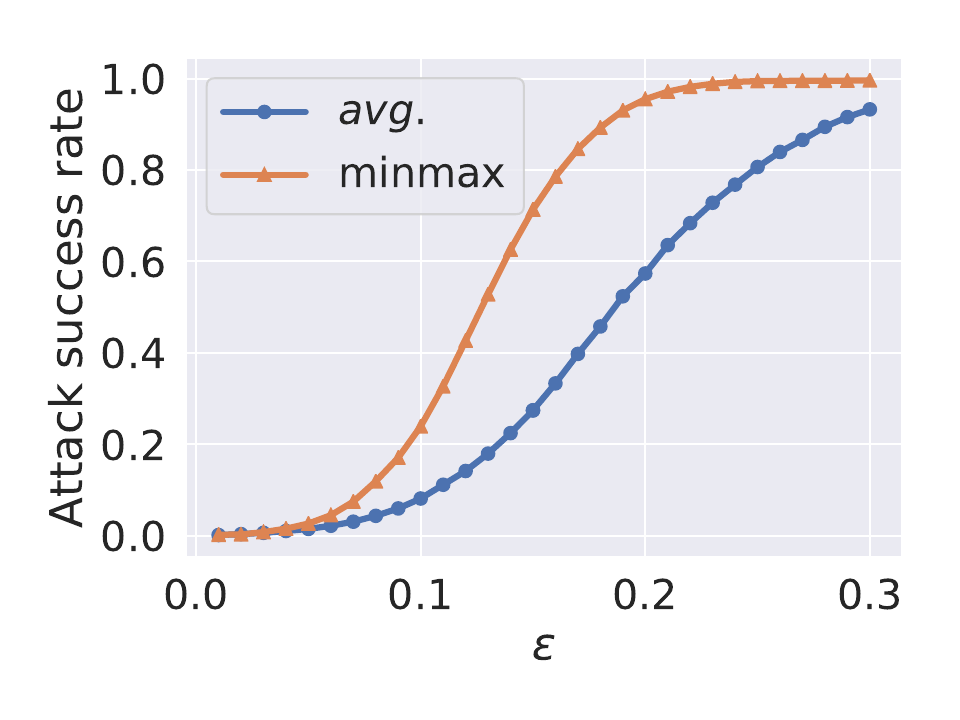} \hspace*{-0.20in} &
\includegraphics[width=0.35\textwidth]{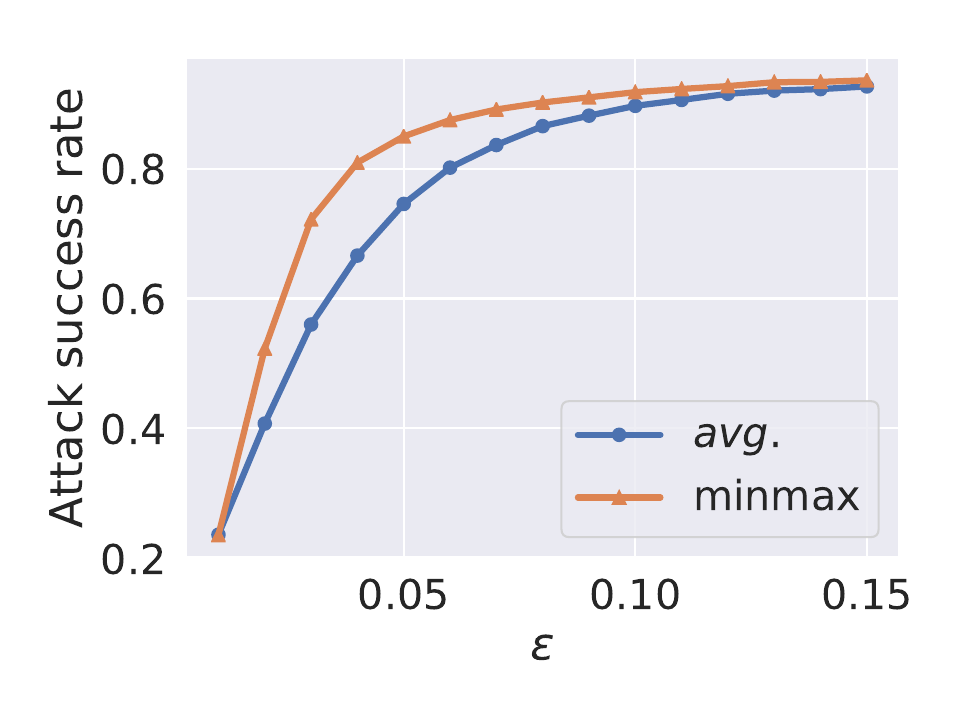} \hspace*{-0.20in} \\
(a) MNIST~\{A, B, C\} \hspace*{-0.20in} & (b) CIFAR-10~\{A, B, C\} %
\end{tabular} 
}
\caption{ASR of 
average and min-max $\ell_\infty$ ensemble attack versus maximum perturbation magnitude $\epsilon$.}
\label{fig:ensemble_eps}
\end{figure}

To perform a boarder evaluation, we repeat the above experiments ($\ell_\infty$ norm) under different $\epsilon$ in Figure~\ref{fig:ensemble_eps}. The ASR of min-max strategy is consistently better or on part with the average strategy. Moreover, APGDA achieves more significant improvement when moderate $\epsilon$ is chosen: MNIST
($\epsilon\in [0.15,0.25]$)
 and CIFAR-10 ($\epsilon \in [0.03,0.05]$).

\subsection{Robust Adversarial Attack over Data Transformations}
\label{ap:sec_data_trans_exps}

Table~\ref{tab:trans_stoch_cifar} %
compare the performance of average (EOT~\cite{athalye18b}) and min-max (APGDA) strategies. Our approach results in 4.31\% %
averaged lift over four models \{A, B, C, D\} on CIFAR-10 under given stochastic and deterministic transformation sets.

\begin{table}[htbp!]
\vspace{-0.5mm}
\caption{Comparison of average and min-max optimization on robust attack over multiple data transformations on CIFAR-10. Note that all data transformations are conducted stochastically with a probability of 0.8, except for \textit{crop} which randomly crops a central area from original image and re-size it into $32 \times 32$.
The adversarial examples are generated by 20-step $\ell_\infty$-APGDA ($\epsilon=0.03$) with $\alpha=\frac{1}{2}, \beta=\frac{1}{100}$ and $\gamma=10$.}
\label{tab:trans_stoch_cifar}
\centering
\resizebox{0.83\textwidth}{!}{
\begin{tabular}{@{}c|c|ccccc|cc|c@{}}
\toprule
\ \ Model & Opt. & Acc$_{ori}$ & Acc$_{flh}$ & Acc$_{flv}$ & Acc$_{bri}$ & Acc$_{crop}$ & ASR$_{avg}$ & ASR$_{gp}$ & Lift ($\uparrow$) \\ \midrule
\multirow{2}{*}{A} 
& $avg.$ & 11.55 & 21.60 & 13.64 & 12.30 & 22.37 & 83.71 & 55.97 & - \\
& $\min\max$ & 13.06 & 18.90 & 13.43 & 13.90 & 20.27 & 84.09 & \textbf{59.17} & \textbf{5.72\%} \\
  \midrule
\multirow{2}{*}{B} & $avg.$ & 6.74 & 11.55 & 10.33 & 6.59 & 18.21 & 89.32 & 69.52 & - \\
 & $\min\max$ & 8.19 & 11.13 & 10.31 & 8.31 & 16.29 & 89.15 & \textbf{71.18} & \textbf{2.39\%} \\ \midrule
\multirow{2}{*}{C} & $avg.$ & 8.23 & 17.47 & 13.93 & 8.54 & 18.83 & 86.60 & 58.85 & - \\
 & $\min\max$ & 9.68 & 13.45 & 13.41 & 9.95 & 18.23 & 87.06 & \textbf{61.63} & \textbf{4.72\%} \\ \midrule
\multirow{2}{*}{D} & $avg.$ & 8.67 & 19.75 & 11.60 & 8.46 & 19.35 & 86.43 & 60.96 & - \\
 & $\min\max$ & 10.43 & 16.41 & 12.14 & 10.15 & 17.64 & 86.65 & \textbf{63.64} & \textbf{4.40\%} \\ \bottomrule
\end{tabular}
}
\end{table}

\begin{table}[htb!]
\caption[Robust attack over multiple transformations (Sec.~\ref{sec:adv_train})]{Comparison of average and min-max optimization on robust attack over multiple data transformations on CIFAR-10. Here a new rotation (\textit{rot}) transformation is introduced, where images are rotated 30 degrees clockwise. Note that all data transformations are conducted with a probability of 1.0.
The adversarial examples are generated by 20-step $\ell_\infty$-APGDA ($\epsilon=0.03$) with $\alpha=\frac{1}{2}, \beta=\frac{1}{100}$ and $\gamma=10$.}
\label{tab:trans_deter_cifar}
\centering
\resizebox{0.95\textwidth}{!}{
\begin{tabular}{@{}c|c|ccccccc|cc|c@{}}
\toprule
\ \ Model & Opt. & Acc$_{ori}$ & Acc$_{flh}$ & Acc$_{flv}$ & Acc$_{bri}$ & Acc$_{gam}$ & Acc$_{crop}$ & Acc$_{rot}$ & ASR$_{avg}$ & ASR$_{gp}$ & Lift ($\uparrow$) \\ \midrule
\multirow{2}{*}{A} 
& $avg.$ & 11.06 & 22.37 & 14.81 & 12.32 & 10.92 & 20.40 & 15.89 & 84.60 & 49.24 & - \\
& $\min\max$ & 13.51 & 18.84 & 14.03 & 15.20 & 13.00 & 18.03 & 14.79 & 84.66 & \textbf{52.31} & \textbf{6.23\%} \\
  \midrule
\multirow{2}{*}{B} & $avg.$ & 5.55 & 11.96 & 9.97 & 5.63 & 5.94 & 16.42 & 11.47 & 90.44 & 65.18 & - \\
 & $\min\max$ & 6.75 & 9.13 & 10.56 & 6.72 & 7.11 & 12.23 & 10.80 & 90.96 & \textbf{70.38} & \textbf{7.98\%} \\ \midrule
\multirow{2}{*}{C} & $avg.$ & 7.65 & 22.30 & 15.82 & 8.17 & 8.07 & 15.44 & 15.09 & 86.78 & 49.67 & - \\
 & $\min\max$ & 9.05 & 15.10 & 14.57 & 9.57 & 9.31 & 14.11 & 14.23 & 87.72 & \textbf{55.37} & \textbf{11.48\%} \\ \midrule
\multirow{2}{*}{D} & $avg.$ & 8.22 & 20.88 & 13.49 & 7.91 & 8.71 & 16.33 & 14.98 & 87.07 & 53.52 & - \\
 & $\min\max$ & 10.17 & 14.65 & 13.62 & 10.03 & 10.35 & 14.36 & 13.82 & 87.57 & \textbf{57.36} & \textbf{7.17\%} \\ \bottomrule
\end{tabular}
}
\end{table}

\begin{wrapfigure}{R}{0.5\textwidth}
\vspace{-9.0mm}
\begin{minipage}{0.50\textwidth}
\begin{figure}[H]
    \centering
    \includegraphics[width=0.9\textwidth]{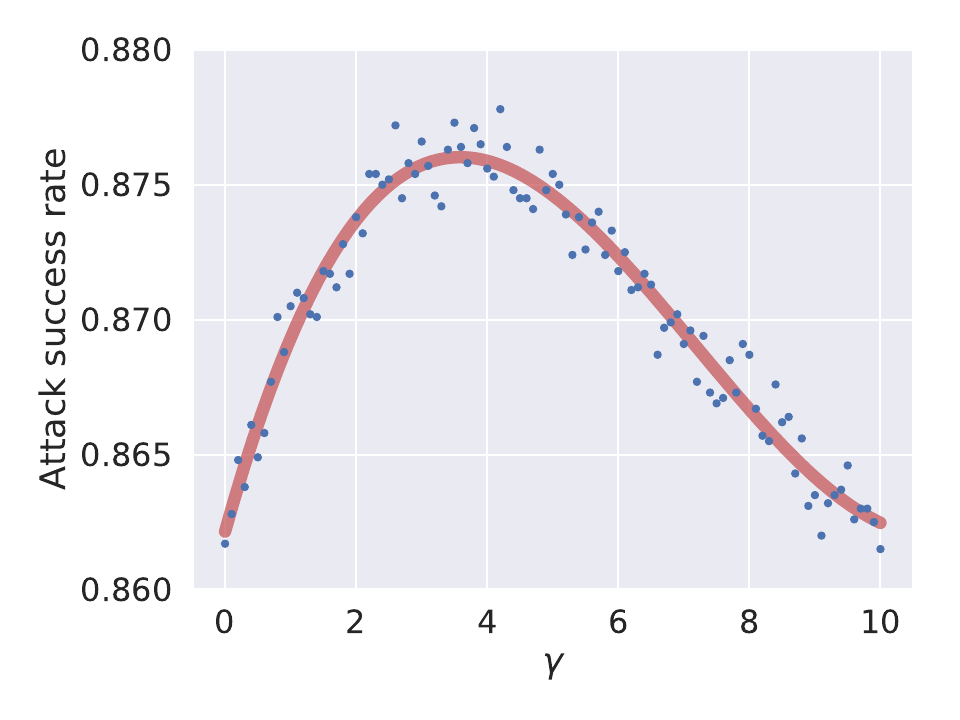}
    \vspace{-5mm}
    \caption{Sensitivity analysis of the regularizer $\frac{\gamma}{2} \| \mathbf w - \mathbf 1/K \|_2^2$ on the probability simplex. The experimental setting is the same as Table~\ref{tab:ensemble_mnist} except for altering the value of $\gamma$.
    }
    \label{fig:regularizer}
\end{figure}
\end{minipage}
\end{wrapfigure}

\subsection{Analysis of Regularization on Probability Simplex}
To further explore the utility of quadratic regularizer on the probability simplex in proposed min-max framework, we conducted sensitivity analysis on $\gamma$ and show how the proposed regularization affects the eventual performance (Figure~\ref{fig:regularizer}a) taking ensemble attack as an example. The experimental setting is the same as Table~\ref{tab:ensemble_mnist} except for altering the value of $\gamma$ from 0 to 10. Figure~\ref{fig:regularizer}a shows that too small or too large $\gamma$ leads to relative weak performance due to the unstable convergence and penalizing too much for average case. When $\gamma$ is around $4$, APGDA will achieve the best performance so we adopted this value in the experiments (Table~\ref{tab:ensemble_mnist}). Moreover, when $\gamma \rightarrow \infty$, the regularizer term dominates the optimization objective and it becomes the average case.

\newpage 
\section{Additional Experiment Results - Adversarial training against multiple types of adversarial attacks}
\label{sec:exp_adv_sub}

\paragraph{Adversarial Training Details:} 
Following the state-of-the-art approach MSD~\cite{msd}, we present experimental results of generalized AT to achieve simultaneous
robustness to $\ell_\infty$, $\ell_2$, and $\ell_1$ perturbations on the MNIST and CIFAR-10 datasets. 
Specifically, we adopted the same architectures as~\cite{msd} four layer convolutional networks on MNIST and the pre-activation version of the ResNet18~\cite{he2016deep}. 
The perturbation radius $\epsilon$ for $(\ell_\infty, \ell_2, \ell_1)$ balls is set as $(0.3, 2.0, 10)$ and $(0.03, 0.5, 12)$ on MNIST and CIFAR-10 following~\cite{msd}. 
For MNIST models, all models are trained 15 epochs with the Adam optimizer. We used a variation of the learning rate schedule from~\cite{smith2018} - piecewise linear schedule from 0 to $10^{-3}$ over the first 6 epochs, and down to 0 over the last 9 epochs.
For CIFAR-10 models, we trained all the models for 50 epochs and used the SGD optimizer with momentum 0.9 and weight decay $5 \times 10^{-4}$. The learning rate schedule rate is piecewise linear from 0 to 0.1 over the first 20 epochs, down to 0.005
over the next 20 epochs, and finally back down to 0 in the last 10 epochs.

\paragraph{Evaluation Setup:} 
To make fair comparisons with MSD~\cite{msd}, we implemented AMPGD based on the public codebase\footnote{\url{https://github.com/locuslab/robust_union}} and followed the exact evaluation settings. Specifically, for $\ell_\infty$ attacks, we use FGSM~\cite{Goodfellow2015explaining}, PGD attack~\cite{madry2017towards} and Momentum Iterative Method~\cite{dong2018mim}. For $\ell_2$ attacks, we use PGD attack, the Gaussian noise attack~\cite{foolbox}, the boundary attack~\cite{brendel2017decision} (Brendel et al., 2017), DeepFool~\cite{moosavi2016deepfool}, the pointwise attack~\cite{SchottRBB19}, DDN-based attack~\cite{RonyHOASG19} and C\&W attack~\cite{carlini2017towards}. For $\ell_1$ attacks, we use the $\ell_1$ PGD attack, the salt \& pepper attack~\cite{foolbox} and the pointwise attack~\cite{SchottRBB19}.
Moreover, we also incorporate the state-of-the-art AutoAttack~\cite{autoattack} for a more comprehensive evaluation under mixed $\ell_p$ perturbations.

\paragraph{Experimental Results:} 
The complete adversarial accuracy results on $\ell_p$ attacks and the union of them are shown in Table~\ref{tab:ampgd_mnist_full}. As we can see, our AMPGD approach leads to a consistent and significant improvement on MNIST. Compared to MSD, we found that our AMPGD emphasize more on defending the strongest adversary - $\ell_\infty$ PGD thus avoiding biased by one particular perturbation model. This observation is also consistent to the learning curves in Figure~\ref{fig:msd_ampgd}.

\begin{table}[htbp!]
\centering
\caption{Summary of adversarial robustness on MNIST.}
\label{tab:ampgd_mnist_full}
\resizebox{0.9\linewidth}{!}{
\begin{tabular}{@{}lccccccc@{}}
\toprule
  &  $L_\infty$-AT & $L_2$-AT & $L_1$-AT  & MAX~\cite{tramer2019adversarial} & AVG~\cite{tramer2019adversarial} & MSD~\cite{msd} & AMPGD           \\ \midrule
Clean Accuracy & 99.1\% & 99.2\% & 99.3\% & 98.6\%      & 99.1\%      & 98.3\%      & 98.3\%          \\ \midrule
$\ell_\infty$ Attacks ($\epsilon=0.3$)~\cite{msd} & 90.3\% & 0.4\% & 0.0\% & 51.0\%      & 65.2\%      & 62.7\%      & 76.1\%          \\
$\ell_2$ Attacks ($\epsilon=2.0$)~\cite{msd} & 13.6\% & 69.2\% & 38.5\% & 61.9\%      & 60.1\%      & 67.9\%      & 70.2\%          \\
$\ell_1$ Attacks ($\epsilon=10$)~\cite{msd} & 4.2\% &  43.4\% & 70.0\% & 52.6\%      & 39.2\%      & 65.0\%      & 67.2\%          \\
All Attacks~\cite{msd} & 3.7\% & 0.4\% & 0.0\% & 42.1\% & 34.9\% & 58.4\% & \textbf{64.1\%} \\ \midrule
AA ($\ell_\infty,\epsilon=0.3$)~\cite{autoattack} & 89.5\% & 0.0\% & 0.0\% & 55.0\% & 52.8\% & 56.6\% & 74.4\% \\ 
AA ($\ell_2,\epsilon=2.0$)~\cite{autoattack} & 3.5\% & 67.6\% & 37.3\% & 56.9\% & 55.8\% & 68.1\% & 63.8\% \\ 
AA ($\ell_1,\epsilon=10$)~\cite{autoattack} & 2.4\% & 60.1\% & 71.9\% & 46.5\% & 40.7\% & 70.0\% & 60.5\% \\
AA (all attacks)~\cite{autoattack} & 1.7\% & 0.0\% & 0.0\% & 36.9\% & 30.5\% & 55.9\% & \textbf{59.3\%} \\\midrule
AA+ ($\ell_\infty,\epsilon=0.3$)~\cite{autoattack} & 89.6\% & 0.0\% & 0.0\% & 54.4\% & 52.4\% & 55.7\% & 74.3\% \\ 
AA+ ($\ell_2,\epsilon=2.0$)~\cite{autoattack} & 2.1\% & 67.4\% & 36.8\% & 55.9\% & 53.8\% & 67.3\% & 61.9\% \\ 
AA+ ($\ell_1,\epsilon=10$)~\cite{autoattack} & 1.8\% & 60.4\% & 71.4\% &  42.3\% & 36.7\% & 68.6\%  & 59.8\% \\
AA+ (all attacks)~\cite{autoattack} & 1.2\% & 0.0\% & 0.0\% & 34.3\% & 28.8\% & 54.8\% & \textbf{58.3\%} \\
\bottomrule
\end{tabular}
}
\end{table}

\newpage 

\section{Interpreting “Image Robustness” with Domain Weights $\mathbf w$}
\label{ap:sec_interpret_w_uni}

Tracking \textit{domain weight} $w$ of the probability simplex from our algorithms is an exclusive feature of solving problem~\ref{eq: prob0}. In Sec.~\ref{sec:experiments}, we show the strength of $w$ in understanding the procedure of optimization and interpreting the adversarial robustness. Here we would like to show the usage of $w$ in measuring ``image robustness'' on devising universal perturbation to multiple input samples. Table~\ref{tab:uni_interpret} and~\ref{tab:uni_interpret2} show the image groups on MNIST with weight $w$ in APGDA and two metrics (distortion of $\ell2$-C\&W, minimum $\epsilon$ for $\ell_\infty$-PGD) of measuring the difficulty of attacking single images. The binary search is utilized to searching for the minimum perturbation.

Although adversaries need to consider a trade-off between multiple images while devising universal perturbation, we find that weighting factor $w$ in APGDA is highly correlated under different $\ell_p$ norms. Furthermore, $w$ is also highly related to minimum distortion required for attacking a single image successfully. It means the inherent ``image robustness'' exists and effects the behavior of generating universal perturbation. Larger weight $w$ usually indicates an image with higher robustness (e.g., fifth 'zero' in the first row of Table~\ref{tab:uni_interpret}), which usually corresponds to the MNIST letter with clear appearance (e.g., bold letter).

\newcolumntype{M}[1]{>{\centering\arraybackslash}m{#1}}
\begin{table}[htb]
\vspace{-1mm}
\caption[Interpretability of domain weight $w$ for universal perturbation (digits 0-4, Sec.~\ref{sec:robust_attack})]{
Interpretability of domain weight $w$ for universal perturbation to multiple inputs on MNIST (\textit{Digit 0 to 4}). Domain weight $w$ for different images under $\ell_p$-norm ($p = 0, 1, 2, \infty$) and two metrics measuring the difficulty of attacking single image are recorded, where dist. ($\ell_2$) denotes the the minimum distortion of successfully attacking images using C\&W ($\ell_2$) attack; $\epsilon_{\min}$ ($\ell_\infty$) denotes the minimum perturbation magnitude for $\ell_\infty$-PGD attack.
}
\label{tab:uni_interpret}
\centering
\setlength\tabcolsep{3pt}
\resizebox{0.90\textwidth}{!}{
\begin{tabular}{@{}c|c|M{1.0cm} M{1.0cm} M{1.0cm} M{1.0cm} M{1.0cm}|
                       M{1.0cm} M{1.0cm} M{1.0cm} M{1.0cm} M{1.0cm}@{}}
\toprule
\multicolumn{2}{c|}{Image} & 
\includegraphics[height=0.39in]{mnist/0/0.jpg} & \includegraphics[height=0.39in]{mnist/0/1.jpg} & \includegraphics[height=0.39in]{mnist/0/2.jpg} & \includegraphics[height=0.39in]{mnist/0/3.jpg} & \includegraphics[height=0.39in]{mnist/0/4.jpg} & \includegraphics[height=0.39in]{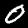} & \includegraphics[height=0.39in]{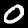} & \includegraphics[height=0.39in]{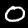} & \includegraphics[height=0.39in]{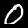} & \includegraphics[height=0.39in]{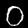}\\ \midrule
\multirow{4}{*}{Weight} & $\ell_0$ & 0. & 0. & 0. & 0. & 1.000 & 0.248 & 0.655 & 0.097 & 0. & 0. \\
 & $\ell_1$ & 0. & 0. & 0. & 0. & 1.000 & 0.07 & 0.922 & 0. & 0. & 0. \\
 & $\ell_2$ & 0. & 0. & 0. & 0. & 1.000 & 0.441 & 0.248 & 0.156 & 0.155 & 0. \\
 & $\ell_\infty$ & 0. & 0. & 0. & 0. & 1.000 & 0.479 & 0.208 & 0.145 & 0.168 & 0. \\ \midrule
\multirow{2}{*}{Metric} & dist.(C\&W $\ell_2$) & 1.839 & 1.954 & 1.347 & 1.698 & 3.041 & 1.545 & 1.982 & 2.178 & 2.349 & 1.050 \\
 & $\epsilon_{\min}$ ($\ell_\infty$) & 0.113 & 0.167 & 0.073 & 0.121 & 0.199 & 0.167 & 0.157 & 0.113 & 0.114 & 0.093 \\ \midrule
 
 \multicolumn{2}{c|}{Image} & 
\includegraphics[height=0.39in]{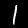} & \includegraphics[height=0.39in]{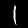} & \includegraphics[height=0.39in]{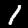} & \includegraphics[height=0.39in]{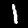} & \includegraphics[height=0.39in]{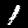} & \includegraphics[height=0.39in]{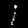} & \includegraphics[height=0.39in]{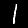} & \includegraphics[height=0.39in]{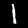} & \includegraphics[height=0.39in]{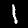} & \includegraphics[height=0.39in]{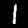}\\ \midrule
\multirow{4}{*}{Weight} & $\ell_0$ & 0. & 0. & 0.613 & 0.180 & 0.206 & 0. & 0. & 0.223 & 0.440 & 0.337 \\
 & $\ell_1$ & 0. & 0. & 0.298 & 0.376 & 0.327 & 0. & 0. & 0.397 & 0.433 & 0.169 \\
 & $\ell_2$ & 0. & 0. & 0.387 & 0.367 & 0.246 & 0. & 0.242 & 0.310 & 0.195 & 0.253 \\
 & $\ell_\infty$ & 0.087 & 0.142 & 0.277 & 0.247 & 0.246 & 0. & 0.342 & 0.001 & 0.144 & 0.514 \\ \midrule
\multirow{2}{*}{Metric} & dist.(C\&W $\ell_2$) & 1.090 & 1.182 & 1.327 & 1.458 & 0.943 & 0.113 & 1.113 & 1.357 & 1.474 & 1.197 \\
 & $\epsilon_{\min}$ ($\ell_\infty$) & 0.075 & 0.068 & 0.091 & 0.105 & 0.096 & 0.015 & 0.090 & 0.076 & 0.095 & 0.106 \\ \midrule
 
\multicolumn{2}{c|}{Image} & 
\includegraphics[height=0.39in]{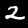} & \includegraphics[height=0.39in]{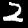} & \includegraphics[height=0.39in]{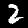} & \includegraphics[height=0.39in]{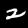} & \includegraphics[height=0.39in]{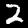} & \includegraphics[height=0.39in]{mnist/2_new/0.jpg} & \includegraphics[height=0.39in]{mnist/2_new/1.jpg} & \includegraphics[height=0.39in]{mnist/2_new/2.jpg} & \includegraphics[height=0.39in]{mnist/2_new/3.jpg} & \includegraphics[height=0.39in]{mnist/2_new/4.jpg}\\ \midrule
\multirow{4}{*}{Weight} & $\ell_0$ & 0. & 1.000 & 0. & 0. & 0. & 0. & 0. & 0.909 & 0. & 0.091 \\
 & $\ell_1$ & 0. & 1.000 & 0. & 0. & 0. & 0. & 0. & 0.843 & 0. & 0.157 \\
 & $\ell_2$ & 0. & 0.892 & 0. & 0. & 0.108 & 0. & 0. & 0.788 & 0. & 0.112 \\
 & $\ell_\infty$ & 0. & 0.938 & 0. & 0. & 0.062 & 0. & 0. & 0.850 & 0. & 0.150 \\ \midrule
\multirow{2}{*}{Metric} & dist.(C\&W $\ell_2$) & 1.335 & 2.552 & 2.282 & 1.229 & 1.884 & 1.928 & 1.439 & 2.312 & 1.521 & 2.356 \\
 & $\epsilon_{\min}$ ($\ell_\infty$)  & 0.050 & 0.165 & 0.110 & 0.083 & 0.162 & 0.082 & 0.106 & 0.176 & 0.072 & 0.171 \\ \midrule
 
\multicolumn{2}{c|}{Image} & 
\includegraphics[height=0.39in]{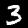} & \includegraphics[height=0.39in]{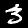} & \includegraphics[height=0.39in]{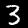} & \includegraphics[height=0.39in]{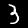} & \includegraphics[height=0.39in]{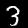} & \includegraphics[height=0.39in]{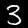} & \includegraphics[height=0.39in]{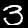} & \includegraphics[height=0.39in]{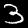} & \includegraphics[height=0.39in]{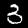} & \includegraphics[height=0.39in]{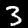}\\ \midrule
\multirow{4}{*}{Weight} & $\ell_0$ & 0.481 & 0. & 0.378 & 0. & 0. & 0. & 0.352 & 0. & 0. & 0.648 \\
 & $\ell_1$ & 0.690 & 0. & 0.310 & 0. & 0. & 0. & 0.093 & 0.205 & 0. & 0.701 \\
 & $\ell_2$ & 0.589 & 0.069 & 0.208 & 0. & 0.134 & 0.064 & 0.260 & 0.077 & 0. & 0.600 \\
 & $\ell_\infty$ & 0.864 & 0. & 0.084 & 0. & 0.052 & 0.079 & 0.251 & 0.156 & 0. & 0.514 \\ \midrule
\multirow{2}{*}{Metric} & dist.(C\&W $\ell_2$) & 2.267 & 1.656 & 2.053 & 1.359 & 0.861 & 1.733 & 1.967 & 1.741 & 1.031 & 2.413 \\
 & $\epsilon_{\min}$ ($\ell_\infty$)  & 0.171 & 0.088 & 0.143 & 0.117 & 0.086 & 0.100 & 0.097 & 0.096 & 0.038 & 0.132 \\ \midrule
 
\multicolumn{2}{c|}{Image} & 
\includegraphics[height=0.39in]{mnist/4/0.jpg} & \includegraphics[height=0.39in]{mnist/4/1.jpg} & \includegraphics[height=0.39in]{mnist/4/2.jpg} & \includegraphics[height=0.39in]{mnist/4/3.jpg} & \includegraphics[height=0.39in]{mnist/4/4.jpg} & \includegraphics[height=0.39in]{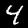} & \includegraphics[height=0.39in]{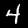} & \includegraphics[height=0.39in]{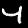} & \includegraphics[height=0.39in]{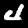} & \includegraphics[height=0.39in]{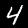}\\ \midrule
\multirow{4}{*}{Weight} & $\ell_0$ & 0. & 0. & 0.753 & 0. & 0.247 & 0. & 0. & 0. & 1.000 & 0. \\
 & $\ell_1$ & 0.018 & 0. & 0.567 & 0. & 0.416 & 0.347 & 0. & 0. & 0.589 & 0.063 \\
 & $\ell_2$ & 0. & 0. & 0.595 & 0. & 0.405 & 0.346 & 0. & 0. & 0.654 & 0. \\
 & $\ell_\infty$ & 0. & 0. & 0.651 & 0. & 0.349 & 0.239 & 0. & 0. & 0.761 & 0. \\ \midrule
\multirow{2}{*}{Metric} & dist.(C\&W $\ell_2$) & 1.558 & 1.229 & 1.939 & 0.297 & 1.303 & 0.940 & 1.836 & 1.384 & 1.079 & 2.027 \\
 & $\epsilon_{\min}$ ($\ell_\infty$)  & 0.084 & 0.088 & 0.122 & 0.060 & 0.094 & 0.115 & 0.103 & 0.047 & 0.125 & 0.100 \\
 \bottomrule
\end{tabular}
}
\end{table}

\begin{table}[htb]
\caption{Interpretability of domain weight $w$ for universal perturbation to multiple inputs on MNIST (\textit{Digit 5 to 9}). Domain weight $w$ for different images under $\ell_p$-norm ($p = 0, 1, 2, \infty$) and two metrics measuring the difficulty of attacking single image are recorded, where dist. ($\ell_2$) denotes the the minimum distortion of successfully attacking images using C\&W ($\ell_2$) attack; $\epsilon_{\min}$ ($\ell_\infty$) denotes the minimum perturbation magnitude for $\ell_\infty$-PGD attack.
}
\label{tab:uni_interpret2}
\centering
\setlength\tabcolsep{3pt}
\resizebox{0.90\textwidth}{!}{
\begin{tabular}{@{}c|c|M{1.0cm} M{1.0cm} M{1.0cm} M{1.0cm} M{1.0cm}|
                       M{1.0cm} M{1.0cm} M{1.0cm} M{1.0cm} M{1.0cm}@{}}
\toprule
\multicolumn{2}{c|}{Image} & 
\includegraphics[height=0.39in]{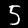} & \includegraphics[height=0.39in]{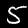} & \includegraphics[height=0.39in]{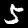} & \includegraphics[height=0.39in]{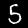} & \includegraphics[height=0.39in]{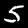} & \includegraphics[height=0.39in]{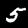} & \includegraphics[height=0.39in]{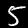} & \includegraphics[height=0.39in]{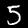} & \includegraphics[height=0.39in]{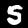} & \includegraphics[height=0.39in]{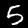}\\ \midrule
\multirow{4}{*}{Weight} & $\ell_0$ & 0. & 0.062 & 0.254 & 0. & 0.684 & 0.457 & 0. & 0. & 0.542 & 0. \\
 & $\ell_1$ & 0.131 & 0.250 & 0. & 0. & 0.619 & 0.033 & 0.157 & 0.005 & 0.647 & 0.158 \\
 & $\ell_2$ & 0.012 & 0.164 & 0.121 & 0. & 0.703 & 0.161 & 0.194 & 0. & 0.508 & 0.136 \\
 & $\ell_\infty$ & 0.158 & 0.008 & 0.258 & 0. & 0.576 & 0.229 & 0.179 & 0. & 0.401 & 0.191 \\ \midrule
\multirow{2}{*}{Metric} & dist. ($\ell_2$) & 1.024 & 1.532 & 1.511 & 1.351 & 1.584 & 1.319 & 1.908 & 1.020 & 1.402 & 1.372 \\
 & $\epsilon_{\min}$ ($\ell_\infty$) & 0.090 & 0.106 & 0.085 & 0.069 & 0.144 & 0.106 & 0.099 & 0.0748 & 0.131 & 0.071 \\ \midrule
 
 \multicolumn{2}{c|}{Image} & 
\includegraphics[height=0.39in]{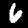} & \includegraphics[height=0.39in]{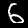} & \includegraphics[height=0.39in]{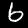} & \includegraphics[height=0.39in]{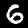} & \includegraphics[height=0.39in]{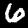} & \includegraphics[height=0.39in]{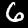} & \includegraphics[height=0.39in]{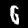} & \includegraphics[height=0.39in]{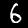} & \includegraphics[height=0.39in]{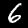} & \includegraphics[height=0.39in]{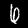}\\ \midrule
\multirow{4}{*}{Weight} & $\ell_0$ & 0.215 & 0. & 0. & 0.194 & 0.590 & 0.805 & 0. & 0. & 0.195 & 0. \\
 & $\ell_1$ & 0.013 & 0. & 0. & 0.441 & 0.546 & 0.775 & 0. & 0. & 0.225 & 0. \\
 & $\ell_2$ & 0.031 & 0. & 0. & 0.410 & 0.560 & 0.767 & 0. & 0. & 0.233 & 0. \\
 & $\ell_\infty$ & 0. & 0. & 0. & 0.459 & 0.541 & 0.854 & 0. & 0. & 0.146 & 0. \\ \midrule
\multirow{2}{*}{Metric} & dist. ($\ell_2$) & 1.199 & 0.653 & 1.654 & 1.156 & 1.612 & 2.158 & 0. & 1.063 & 1.545 & 0.147 \\
 & $\epsilon_{\min}$ ($\ell_\infty$) & 0.090 & 0.017 & 0.053 & 0.112 & 0.158 & 0.159 & 0.020 & 0.069 & 0.145 & 0.134 \\ \midrule
 
\multicolumn{2}{c|}{Image} & 
\includegraphics[height=0.39in]{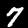} & \includegraphics[height=0.39in]{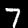} & \includegraphics[height=0.39in]{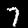} & \includegraphics[height=0.39in]{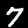} & \includegraphics[height=0.39in]{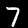} & \includegraphics[height=0.39in]{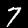} & \includegraphics[height=0.39in]{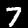} & \includegraphics[height=0.39in]{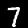} & \includegraphics[height=0.39in]{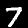} & \includegraphics[height=0.39in]{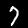}\\ \midrule
\multirow{4}{*}{Weight} & $\ell_0$ & 0.489 & 0. & 0. & 0.212 & 0.298 & 0.007 & 0.258 & 0.117 & 0.482 & 0.136 \\
 & $\ell_1$ & 0.525 & 0.190 & 0. & 0.215 & 0.070 & 0.470 & 0.050 & 0.100 & 0.343 & 0.038 \\
 & $\ell_2$ & 0.488 & 0.165 & 0. & 0.175 & 0.172 & 0.200 & 0.175 & 0.233 & 0.378 & 0.014 \\
 & $\ell_\infty$ & 0.178 & 0.263 & 0. & 0.354 & 0.205 & 0.258 & 0.207 & 0.109 & 0.426 & 0. \\ \midrule
\multirow{2}{*}{Metric} & dist. ($\ell_2$) & 1.508 & 1.731 & 1.291 & 1.874 & 1.536 & 1.719 & 2.038 & 1.417 & 2.169 & 0.848 \\
 & $\epsilon_{\min}$ ($\ell_\infty$)  & 0.110 & 0.125 & 0.089 & 0.126 & 0.095 & 0.087 & 0.097 & 0.084 & 0.135 & 0.077 \\ \midrule
 
\multicolumn{2}{c|}{Image} & 
\includegraphics[height=0.39in]{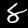} & \includegraphics[height=0.39in]{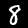} & \includegraphics[height=0.39in]{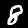} & \includegraphics[height=0.39in]{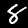} & \includegraphics[height=0.39in]{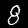} & \includegraphics[height=0.39in]{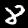} & \includegraphics[height=0.39in]{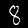} & \includegraphics[height=0.39in]{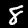} & \includegraphics[height=0.39in]{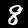} & \includegraphics[height=0.39in]{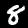}\\ \midrule
\multirow{4}{*}{Weight} & $\ell_0$ & 0. & 0. & 1.000 & 0. & 0. & 0.246 & 0. & 0. & 0. & 0.754 \\
 & $\ell_1$ & 0. & 0.180 & 0.442 & 0.378 & 0. & 0.171 & 0. & 0. & 0. & 0.829 \\
 & $\ell_2$ & 0. & 0.298 & 0.593 & 0.109 & 0. & 0.330 & 0. & 0. & 0. & 0.670 \\
 & $\ell_\infty$ & 0. & 0.377 & 0.595 & 0.028 & 0. & 0.407 & 0. & 0. & 0. & 0.593 \\ \midrule
\multirow{2}{*}{Metric} & dist. ($\ell_2$) & 1.626 & 1.497 & 1.501 & 1.824 & 0.728 & 1.928 & 1.014 & 1.500 & 1.991 & 1.400 \\
 & $\epsilon_{\min}$ ($\ell_\infty$)  & 0.070 & 0.153 & 0.156 & 0.156 & 0.055 & 0.171 & 0.035 & 0.090 & 0.170 & 0.161 \\ \midrule
 
\multicolumn{2}{c|}{Image} & 
\includegraphics[height=0.39in]{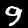} & \includegraphics[height=0.39in]{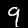} & \includegraphics[height=0.39in]{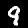} & \includegraphics[height=0.39in]{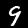} & \includegraphics[height=0.39in]{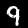} & \includegraphics[height=0.39in]{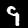} & \includegraphics[height=0.39in]{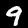} & \includegraphics[height=0.39in]{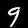} & \includegraphics[height=0.39in]{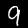} & \includegraphics[height=0.39in]{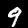}\\ \midrule
\multirow{4}{*}{Weight} & $\ell_0$ & 1. & 0. & 0. & 0. & 0. & 0. & 0.665 & 0.331 & 0. & 0.004 \\
 & $\ell_1$ & 0.918 & 0. & 0.012 & 0. & 0.070 & 0. & 0.510 & 0.490 & 0. & 0. \\
 & $\ell_2$ & 0.911 & 0. & 0.089 & 0. & 0. & 0. & 0.510 & 0.490 & 0. & 0. \\
 & $\ell_\infty$ & 0.935 & 0. & 0.065 & 0. & 0. & 0. & 0.665 & 0.331 & 0. & 0.004 \\ \midrule
\multirow{2}{*}{Metric} & dist. ($\ell_2$) & 1.961 & 1.113 & 1.132 & 1.802 & 0.939 & 1.132 & 1.508 & 1.335 & 1.033 & 1.110 \\
 & $\epsilon_{\min}$ ($\ell_\infty$)  & 0.144 & 0.108 & 0.083 & 0.103 & 0.079 & 0.041 & 0.090 & 0.103 & 0.083 & 0.044 \\
 \bottomrule
\end{tabular}
}
\end{table}